\documentclass[10pt,journal,compsoc]{IEEEtran}

\ifCLASSOPTIONcompsoc
  \usepackage[nocompress]{cite}
\else
  \usepackage{cite}
\fi

\ifCLASSINFOpdf
\else
\fi

\usepackage{graphicx}
\usepackage{amsmath}
\usepackage{amssymb}
\usepackage{booktabs}
\usepackage{bm}

\usepackage{subcaption}
\usepackage{booktabs}
\usepackage{multirow}
\usepackage{tabularx}
\usepackage{makecell}
\usepackage{soul}  %
\usepackage{float}

\usepackage[pagebackref,breaklinks,colorlinks]{hyperref}

\usepackage{xspace}
\makeatletter
\DeclareRobustCommand\onedot{\futurelet\@let@token\@onedot}
\def\@onedot{\ifx\@let@token.\else.\null\fi\xspace}

\def\eg{\emph{e.g}\onedot} 
\def\ie{\emph{i.e}\onedot} 
 
\def\etc{\emph{etc}\onedot} \def\vs{\emph{vs}\onedot}

\makeatother

\newcommand{\red}[1]{{{#1}}} %

\begin{document}
\title{

Unifying Flow, Stereo and Depth Estimation

}

\author{Haofei~Xu,
Jing~Zhang,
Jianfei~Cai,~\IEEEmembership{Fellow,~IEEE,}
Hamid~Rezatofighi,
Fisher~Yu,
Dacheng~Tao,~\IEEEmembership{Fellow,~IEEE,}
and~Andreas~Geiger
\IEEEcompsocitemizethanks{
\IEEEcompsocthanksitem H. Xu is with ETH Zurich, Switzerland and University of T\"ubingen, Germany. Email: haofei.xu@vision.ee.ethz.ch
\IEEEcompsocthanksitem J. Zhang and D. Tao are with the Sydney AI Center and The University of Sydney, Australia. Email: jing.zhang1@sydney.edu.au, dacheng.tao@sydney.edu.au
\IEEEcompsocthanksitem J. Cai and H. Rezatofighi are with Monash University, Australia. Email: jianfei.cai@monash.edu, hamid.rezatofighi@monash.edu
\IEEEcompsocthanksitem F. Yu is with ETH Zurich, Switzerland. Email: i@yf.io
\IEEEcompsocthanksitem A. Geiger is with University of T\"ubingen and Max Planck Institute for Intelligent Systems, Germany. Email: a.geiger@uni-tuebingen.de
}
}

\IEEEtitleabstractindextext{%
\begin{abstract}
We present a unified formulation and model for three motion and 3D perception tasks: optical flow, rectified stereo matching and unrectified stereo depth estimation from posed images. Unlike previous specialized architectures for each specific task, we formulate all three tasks as a unified dense correspondence matching problem, which can be solved with a single model by directly comparing feature similarities. Such a formulation calls for discriminative feature representations, which we achieve using a Transformer, in particular the cross-attention mechanism. We demonstrate that cross-attention enables integration of knowledge from another image via cross-view interactions, which greatly improves the quality of the extracted features. Our unified model naturally enables cross-task transfer since the model architecture and parameters are shared across tasks. We outperform RAFT with our unified model on the challenging Sintel dataset, and our final model that uses a few additional task-specific refinement steps outperforms or compares favorably to recent state-of-the-art methods on 10 popular flow, stereo and depth datasets, while being simpler and more efficient in terms of model design and inference speed.

\end{abstract}

}

\maketitle

\IEEEdisplaynontitleabstractindextext

\IEEEpeerreviewmaketitle

\IEEEraisesectionheading{\section{Introduction}\label{sec:introduction}}

\IEEEPARstart{U}{nderstanding} the 3D scene structure and motion from a set of 2D images has been a long-standing goal of computer vision~\cite{hartley2003multiple,sun2010secrets}. It is the cornerstone of many real-world applications, such as reconstructing a 3D city from internet photos~\cite{agarwal2011building}, action recognition with optical flow~\cite{simonyan2014two}, augmented reality~\cite{klein2007parallel} and autonomous driving~\cite{geiger2012we}.

Classic approaches typically tackle these tasks by solving an energy minimization problem with optimization techniques. For example, the variational approach for optical flow~\cite{horn1981determining}, semi-global matching for stereo vision~\cite{hirschmuller2007stereo} and bundle adjustment for structure-from-motion~\cite{triggs1999bundle}. Although significant progress has been made with classic methods, they often still struggle in challenging situations like textureless regions and thin structures.

\begin{figure}[t]
    \centering
    \includegraphics[width=0.95\linewidth]{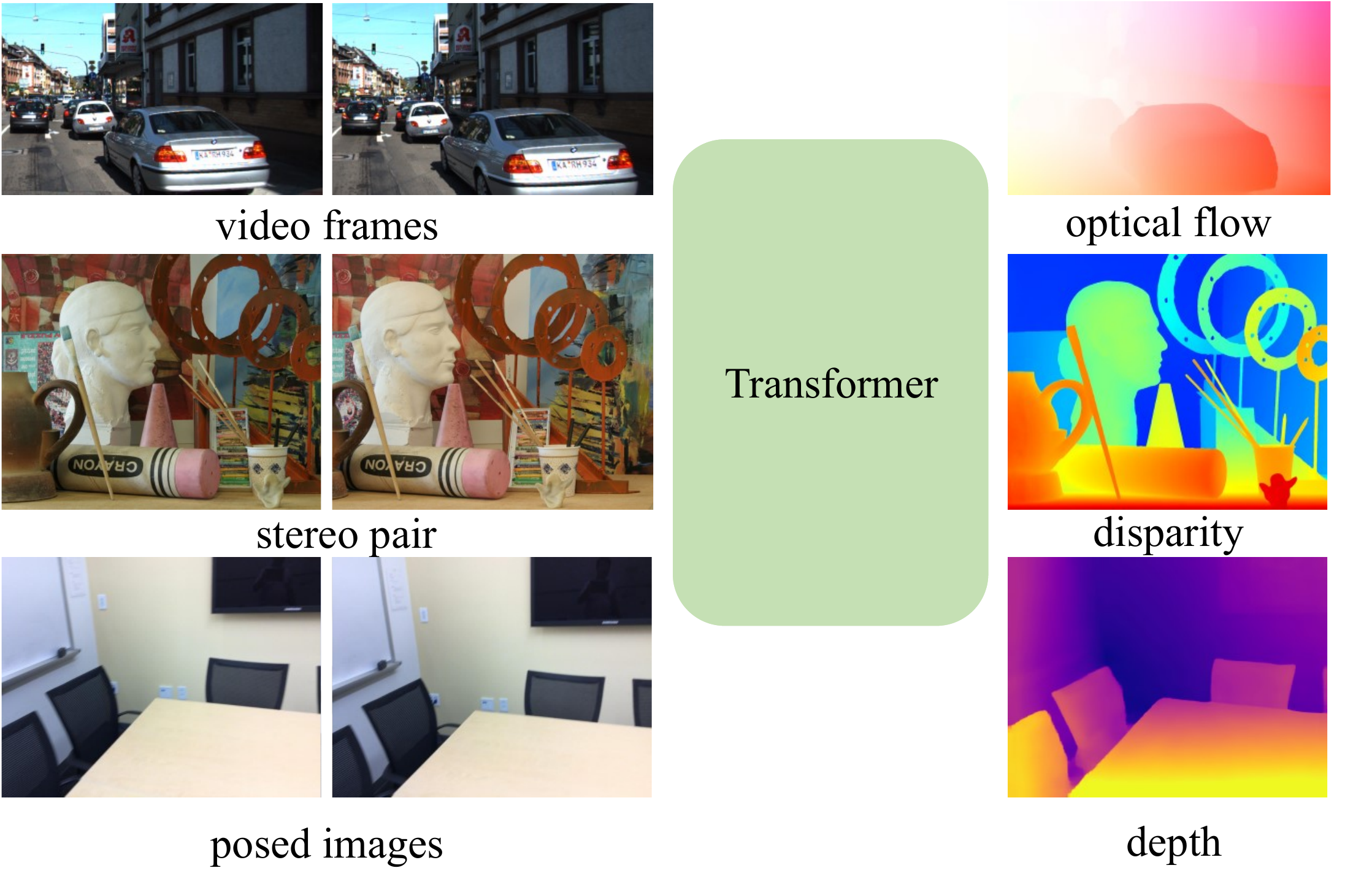}
    \caption{\textbf{A unified model for three motion and 3D perception tasks}. We formulate optical flow, rectified stereo matching and unrectified stereo depth estimation as a unified dense correspondence matching problem that can be solved by directly comparing feature similarities. To obtain discriminative features for matching, we use a Transformer, in particular the cross-attention mechanism. We demonstrate that cross-attention can integrate the knowledge from another view via cross-view interactions, which greatly improves the quality of the extracted features. This is not achievable with classical convolution-based backbones which operate on each view independently.
    }
    \label{fig:teaser}
    \vspace{-10pt}
\end{figure}

The rapid advancement of deep learning~\cite{krizhevsky2012imagenet} and large-scale datasets also enables direct feed-forward inference of geometry and motion using high-capacity deep neural networks. Different network architectures have been proposed for different tasks in the last few years (\eg, FlowNet~\cite{dosovitskiy2015flownet} for optical flow and MVSNet~\cite{yao2018mvsnet} for multi-view stereo). Further development of network architectures has led to steady progress on geometry and motion tasks, and learning-based methods are currently dominating the leaderboards of popular benchmarks~\cite{butler2012naturalistic,geiger2012we,menze2015object,scharstein2014high}.

However, existing works are largely driven by designing task-specific models to solve each task independently, and thus a large variety of network architectures~\cite{ilg2017flownet,sun2018pwc,yin2019hierarchical,xu2020aanet,Tang2019BANetDB,teed2020raft,xu2022attention} have been proposed to handle different tasks, ignoring the fact that many multi-view geometry and motion tasks are fundamentally related correspondence estimation problems. Such a task-specific design philosophy inevitably leads to lots of architectures to deal with, and additional complexities are introduced in model deployment or update for real-world applications. Besides, pretrained models for different tasks cannot be reused (\eg, transfer between tasks) when they are studied in isolation.

In this paper, we aim at developing a single unified model to solve three dense perception tasks: optical flow, rectified stereo matching and unrectified stereo depth estimation from posed images, as shown in Fig.~\ref{fig:teaser}, which are fundamental building blocks for motion (optical flow) and 3D (depth) understanding. To achieve this, we first identify the main obstacle that hinders previous models to be generally applicable. In particular, previous methods mostly encode the task-specific geometric inductive bias (\eg, the cost volume~\cite{scharstein2002taxonomy,sun2018pwc} with different shapes) as intermediate components of the model and use subsequent convolutional networks for flow/disparity/depth regression. Since the geometric inductive bias is task-dependent (\eg, optical flow's cost volume is typically based on 2D correlation~\cite{dosovitskiy2015flownet}, while stereo matching networks construct cost volume by 1D correlation~\cite{xu2020aanet} or feature concatenation~\cite{chang2018pyramid}), this leads to task-specific convolutional architectures for post-processing the cost volume. Moreover, the type of convolutional networks can be quite different (2D~\cite{sun2018pwc,xu2020aanet}, 3D~\cite{chang2018pyramid,zhang2019ga} or ConvGRU~\cite{teed2020raft,lipson2021raft}), which introduces additional challenges in unifying these tasks under such a pipeline.

Our key insight is that these tasks can be unified in an explicit dense correspondence matching formulation, where they can be solved by directly comparing feature similarities. Thus the task is reduced to learning strong task-agnostic feature representations for matching, for which we use a Transformer~\cite{vaswani2017attention}, in particular the cross-attention mechanism to achieve this. We demonstrate that cross-attention can integrate the knowledge from another image via cross-view interactions, which greatly improves the quality of the extracted features. In our method, the geometric inductive biases for each task are modeled with \textit{parameter-free} task-specific matching layers at the final output, which not only introduces no task-specific learnable parameters, but also demonstrates that cost volume post-processing is not always necessary for geometry and motion estimation tasks once we have strong features. This is different from Perceiver IO~\cite{jaegle2021perceiver} that directly regresses optical flow without considering any geometric inductive bias, which is less efficient in terms of model parameters (ours is $8\times$ less) and inference speed (ours is $4\times$ faster). It also differs from IIB~\cite{yifan2022input} that injects the geometric inductive bias at the input, which makes subsequent network layers task-specific.  Our formulation implicitly assumes the corresponding pixels are visible on both images and thus they can be matched by comparing feature similarities. To handle unmatched (occluded and out-of-boundary) regions, we introduce a simple task-agnostic self-attention layer to propagate the high-quality predictions to unmatched regions by measuring feature self-similarity~\cite{hui2020liteflownet3,Jiang_2021_ICCV}.

Our unified model naturally enables cross-task transfer since each task uses exactly the same learnable parameters for feature extraction. For example, without any finetuning, a pretrained optical flow model can be directly used for the task of rectified stereo matching and unrectified stereo depth estimation. Moreover, when finetuning with the pretrained flow model as initialization, we not only enjoy faster training speed for stereo and depth, but also achieve better performance, as evidenced by our experiments (Table~\ref{tab:cross_task_transfer}).

{Our unified model with only one task-agnostic hierarchical matching refinement outperforms RAFT~\cite{teed2020raft} with 31 refinement steps on the challenging Sintel~\cite{butler2012naturalistic} dataset while running faster (Fig.~\ref{fig:flow_iter_vs_epe} and Table~\ref{tab:flow_raft_vs_gmflow}), demonstrating the effectiveness and efficiency of our method. Our final model that uses a few additional task-specific refinement steps outperforms or compares favorably to recent state-of-the-art methods on 10 popular flow/stereo/depth datasets (KITTI Flow~\cite{menze2015object}, Sintel~\cite{butler2012naturalistic}, Middlebury~\cite{scharstein2014high}, KITTI Stereo~\cite{menze2015object}, ETH3D Stereo~\cite{schops2017multi}, Argoverse Stereo~\cite{chang2019argoverse}, ScanNet~\cite{dai2017scannet}, SUN3D~\cite{xiao2013sun3d}, RGBD-SLAM~\cite{sturm2012benchmark} and Scenes11~\cite{ummenhofer2017demon}), while being simpler and more efficient in terms of model design and inference speed.}

This work represents a substantial extension of our previous CVPR 2022 conference paper GMFlow~\cite{xu2022gmflow}, where the new contributions are summarized as follows: (1) The initial work GMFlow~\cite{xu2022gmflow} aims at demonstrating a successful alternative to RAFT's~\cite{teed2020raft} iterative architecture for the optical flow task, while this work proposes a more holistic perspective that unifies three dense correspondence estimation tasks. (2) We extend GMFlow to rectified stereo matching and unrectified stereo depth estimation from posed images and conduct extensive experiments. (3) We study the cross-task transfer behavior by reusing pretrained models. Our project page is available at \href{https://haofeixu.github.io/unimatch}{haofeixu.github.io/unimatch}, and our code and models are available at \href{https://github.com/autonomousvision/unimatch}{github.com/autonomousvision/unimatch}.

\section{Related Work}

Most existing methods for optical flow, rectified stereo matching and unrectified stereo depth estimation have been largely driven by designing specific architectures for each specific task, without pursing a unified model. In this section, we will first review the development of each task independently, and then discuss their relations from the perspective of a unified model and multi-task learning.

\subsection{Optical Flow}

Optical flow has been traditionally tackled with variational approaches~\cite{horn1981determining,bruhn2005lucas,brox2010large,brox2004high,xu2011motion,sun2010secrets}, where it is typically solved as an energy minimization problem that consists of a brightness constancy term and a regularization term. The advancement of deep learning has also enabled directly learning optical flow from data. The pioneering learning-based work, FlowNet~\cite{dosovitskiy2015flownet}, proposed a convolutional neural network that directly takes two images as input and regresses an optical flow field. Further advances of network architectures and training strategies~\cite{ilg2017flownet,ranjan2017optical,sun2018pwc,hur2019iterative,teed2020raft,Jiang_2021_ICCV,xu2021high,sun2021autoflow,bai2022deep,luo2022blearning,zhang2021separable,sui2022craft,huang2022flowformer,zheng2022dip,luo2022learning} have led to steady progress for learning-based methods, which today outperform traditional approaches by a large margin and are currently dominating the benchmarks including Sintel~\cite{butler2012naturalistic} and KITTI~\cite{geiger2012we,menze2015object}. 

However, a closer look at existing learning-based approaches reveals that the underlying architectural principles haven't changed much since FlowNet~\cite{dosovitskiy2015flownet}, that is, regressing optical flow from local correlation (\ie, cost volume) with convolutions. Such a local regression approach is intrinsically limited by trading off large-displacement flow estimation with the size of the cost volume. To alleviate this problem, two popular strategies are coarse-to-fine~\cite{ranjan2017optical,sun2018pwc} and iterative refinement~\cite{hur2019iterative,teed2020raft} methods, which estimate large displacements incrementally in multiple stages. However, coarse-to-fine methods tend to miss fast-moving small objects if the resolution is too coarse and may suffer from the error-propagation issue~\cite{revaud2015epicflow}. In contrast, the iterative approaches like RAFT~\cite{teed2020raft} lead to a linear increase in processing time due to the large number of sequential refinements. In contrast, we reformulate optical flow as a global matching problem, which identifies dense correspondences by directly comparing pair-wise feature similarities, leading to significant improvement for large displacements.

\subsection{Stereo Matching}

Typical stereo matching methods generally follow a four-step pipeline~\cite{scharstein2002taxonomy}: matching cost computation, cost aggregation, disparity computation and disparity refinement. Again, early optimization-based methods~\cite{hosni2012fast,yoon2006adaptive} have been replaced by modern deep learning-based approaches~\cite{Zbontar2016Stereo,mayer2016large,kendall2017end,chang2018pyramid,zhang2019ga,xu2020aanet,shen2021cfnet,poggi2021synergies}. The current representative stereo methods can be broadly classified into two categories: 3D and 2D convolution-based approaches. Their key difference lies in the cost volume construction method. 3D convolution-based methods~\cite{chang2018pyramid,zhang2019ga,shen2021cfnet,xu2022attention} typically use feature concatenation while 2D methods~\cite{mayer2016large,xu2020aanet} use feature correlation. These methods usually build a \textit{local} cost volume with a predefined search space (typically 192 pixels~\cite{kendall2017end}) and the final disparity prediction is obtained by computing the weighted sum of all disparity candidates. Thus the output is always constrained by the predefined disparity range, which makes these methods less flexible to handle unconstrained settings like high-resolution images or new camera settings. For example, to adapt such an architecture to larger disparity ranges, the full model has to be re-trained by setting a new predefined maximum disparity. In contrast, we directly perform \textit{global} matching along the scanline, which make no assumption on the disparity range and is able to handle arbitrary image resolutions. 

Recent iterative 2D methods like RAFT-Stereo~\cite{lipson2021raft} and CREStereo~\cite{li2022practical} mostly follow the high-level design of the RAFT~\cite{teed2020raft} architecture for optical flow, while introducing several task-specific components (\eg, 1D correlation) to make such a method suitable for the stereo matching task. In contrast, we show that our matching-based perspective enables to use the \textit{same} model for both optical flow and stereo matching, with exactly the same learnable parameters. Besides, our model is also more efficient since we don't rely on any 3D convolutions or a large number of sequential refinements. On the other hand, although MC-CNN~\cite{Zbontar2016Stereo} also tries to learn strong features for matching, the features in MC-CNN are extracted \textit{independently} with a convolutional network, without considering cross-view interactions. However, as evidenced by our results, cross-view interactions are crucial for strong and discriminative features (see Table~\ref{tab:flow_conv_vs_softmax} and Table~\ref{tab:transformer}). 

Perhaps the most related stereo work to ours is STTR~\cite{li2021revisiting}, which also uses a Transformer and matching-based disparity computation. However, STTR relies on a complex optimal transport matching layer and doesn’t produce predictions for occluded pixels, while we use a much simpler softmax operation and a simple flow propagation layer to handle occlusions. The later CSTR (Context-Enhanced Stereo Transformer)~\cite{guo2022context} tries to improve STTR's performance with a new Transformer architecture, but it still suffers from the limitation of STTR. Moreover, STTR is designed to solve the stereo matching task, while we are seeking a unified model applicable to three different dense correspondence estimation tasks.

\subsection{Depth Estimation}

Learning-based depth estimation methods can be broadly categorized into monocular and multi-view approaches. Monocular methods~\cite{garg2016unsupervised,godard2017unsupervised,zhou2017unsupervised,xu2019rdn4depth,godard2019digging,ranftl2020towards} take a single image as input and use generic network architectures like ResNet~\cite{he2016deep} to predict the dense depth map, while multi-view methods~\cite{ummenhofer2017demon,im2019dpsnet,Tang2019BANetDB,Teed2020DeepV2D,yao2018mvsnet,guizilini2022multi,sayed2022simplerecon,ma2022multiview,guizilini2022depth} usually focus on how to encode the geometric inductive bias (cost volume, warping, \etc) into the network architecture. Compared with monocular methods, multi-view depth estimation can better leverage the information from additional viewpoints and usually lead to improved performance~\cite{watson2021temporal}. Since multi-view information (\eg, video sequences) are usually readily available for many applications, we consider multi-view depth estimation in this paper. A popular multi-view depth pipeline is using the plane-sweep stereo~\cite{im2019dpsnet,gu2020cascade} approach, where different depth planes are tested for correctness. However, like rectified stereo matching, the state-of-the-art methods are usually dominated by 3D convolution-based approaches~\cite{im2019dpsnet,Teed2020DeepV2D}, which accordingly introduces cubic computational complexity.  In this paper, we approach this task from an explicit matching-based perspective and use a Transformer to obtain strong features for matching, achieving highly competitive performance without relying on any 3D convolutions. This is different from the recent work TransMVSNet~\cite{ding2022transmvsnet}, which still relies on 3D convolutions for cost volume post-processing and where the Transformer is used before the cost volume construction stage. Thus, our method is simpler and more lightweight.

\begin{figure*}
    \centering
    \includegraphics[width=\linewidth]{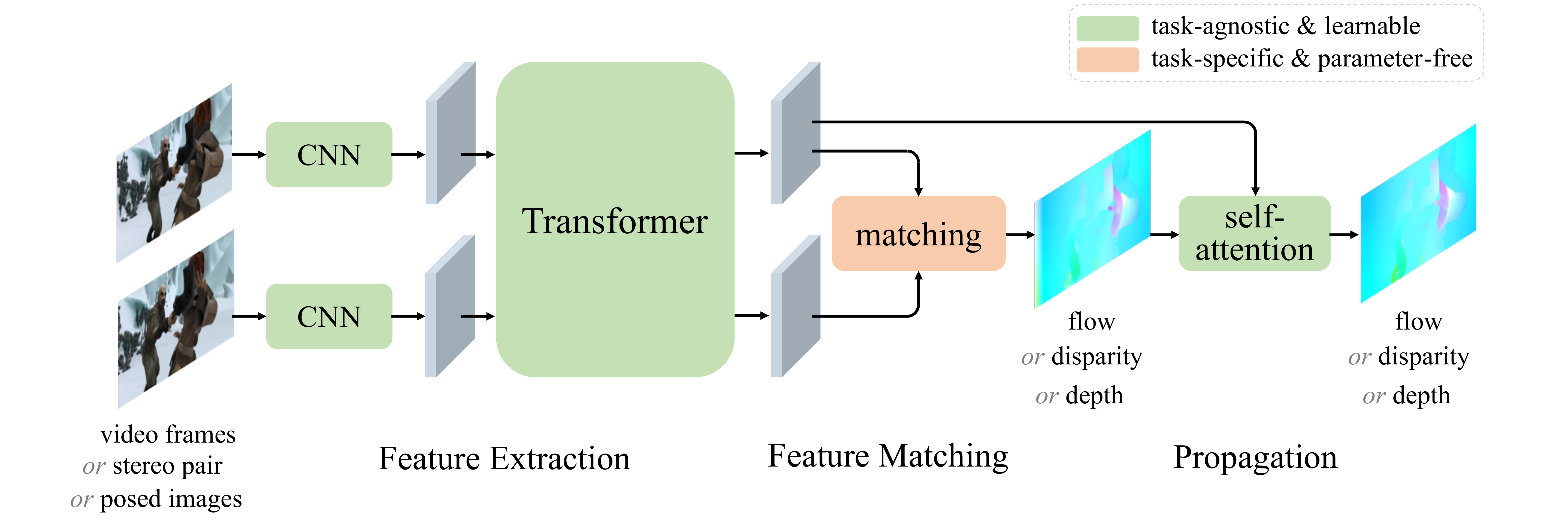}
    \caption{\textbf{Overview of our unified model}. We consider two images as input, which can be video frames for optical flow, a stereo pair for rectified stereo matching or posed images for unrectified stereo depth estimation. We first extract $8\times$ downsampled dense features from the two input images with a weight-sharing convolutional network. The features are then fed into a Transformer for feature enhancement. Next, we perform feature matching by using parameter-free task-specific matching layers, which produce the optical flow, disparity or depth output, depending on the task. An additional task-agnostic self-attention layer is introduced to propagate the high-quality predictions to unmatched regions by measuring feature self-similarity.}
    \label{fig:overview}
    \vspace{-10pt}

\end{figure*}

\subsection{Unified Model}

Unified models aim at using task-agnostic architectures to solve different tasks. One notable work is Perceiver IO~\cite{jaegle2021perceiver}, which proposes a general Transformer architecture for different problems in different domains. Perceiver IO has been applied to the optical flow task, where a direct concatenation of two input images is fed to the Transformer, and optical flow is regressed without using any inductive bias. Despite its architectural simplicity, more parameters ($8\times$ more than ours) and additional computational complexity ($4\times$ slower than ours) are introduced in order to make the model perform well. Perceiver IO has also been used to solve the unrectified stereo depth estimation task~\cite{yifan2022input}, where the geometric inductive bias is fed into the network as additional inputs. Different from Perceiver IO, our design is motivated from a unified perspective that learns strong feature representations for dense correspondence matching for geometry and motion tasks. In our method, the geometric inductive biases are well-preserved at the final \textit{parameter-free} matching layers, which doesn't introduce any task-specific learnable parameters. This is different from Perceier IO for optical flow and stereo depth estimation, where the network inputs are task-specific and thus it is not easy to reuse the model parameters from different tasks. Another related work is HD3~\cite{yin2019hierarchical}, which proposes a model that is applicable to both optical flow and stereo matching. However, HD3 relies on task-specific correlations (2D or 1D) as intermediate network components, resulting to task-specific learnable parameters in the subsequent decoders and thus making it not easy to transfer pretrained models across tasks.

We also note that our unified model is different from multi-task joint training approaches~\cite{jiang2019sense,yin2018geonet,zou2018df,ranjan2019competitive,wang2019unos}. Multi-task methods usually deploy task-specific network architectures which are trained together. In contrast, our unified model focuses on designing a single network that is generally applicable to several tasks, without any task-specific learnable parameters.

\section{Methodology}

Dense correspondences between different viewpoints are the core of optical flow, rectified stereo matching and unrectified stereo depth estimation tasks. To unify these three tasks, our key idea is to use an explicit dense correspondence matching formulation, which identifies the solution by directly comparing feature similarities. Such a formulation calls for discriminative features, for which we use a Transformer, in particular the cross-attention to achieve this. The cross-attention can integrate the knowledge from another image via cross-view interactions, which greatly improves features' quality and is not achievable with convolutions that operate on each view independently~\cite{Zbontar2016Stereo}.

\red{
Fig.~\ref{fig:overview} provides an overview of our proposed method. We first extract dense features from two input images and then obtain the prediction with a parameter-free matching layer. A final self-attention layer is used to propagate the high-quality predictions to unmatched regions by measuring feature self-similarity.
}

In the following, we will first formulate the differentiable matching layers for optical flow, rectified stereo matching and unrectified stereo depth estimation, and then present a unified Transformer-based model to extract strong features for matching. Note that the matching layers are designed by considering different constraints for each task, which are therefore task-specific. However, the matching layers are parameter-free since they only compare feature similarities. The learnable parameters for all three tasks are exactly the same and thus they can be reused for cross-task transfer.

\subsection{Formulation}
\label{sec:formulation}

We consider two images ${\bm I}_1$ and ${\bm I}_2$ as input. These can be video frames for optical flow, a stereo pair for rectified stereo matching, or two posed images (with known camera intrinsic and extrinsic parameters) for unrectified stereo depth estimation. We assume that $8 \times$ downsampled dense features ${\bm F}_1, {\bm F}_2 \in \mathbb{R}^{H \times W \times D}$ are extracted for both images (we will provide details on our feature extractor in Sec.~\ref{sec:feature}), where $H, W$ and $D$ denote height, width and feature dimension, respectively. Next, we present the parameter-free task-specific matching layers for optical flow, rectified stereo matching and unrectified stereo depth estimation under our unified matching-based formulation.

\subsubsection{Flow Matching}
\label{sec:flow_matching}

Optical flow represents the apparent motion between two video frames, which can be computed by finding 2D pixel-wise dense correspondences on the image plane. To achieve this, we directly compare the feature similarities for each location in ${\bm F}_1$ with respect to all locations in ${\bm F}_2$ by computing their correlations (\ie, global matching). This can be implemented efficiently using a simple matrix multiplication:
\begin{equation}
\label{eq:flow_corr}
    {\bm C}_{\mathrm{flow}} =  \frac{{\bm F}_1 {\bm F}_2^T}{\sqrt{D}}  \in \mathbb{R}^{H \times W \times H \times W}, 
\end{equation}
where each element in the correlation matrix ${\bm C}_{\mathrm{flow}}$ represents the correlation value between coordinates ${\bm p}_1 = (i, j)$ in ${\bm F}_1$ and ${\bm p}_2 = (k, l)$ in ${\bm F}_2$, and $\frac{1}{\sqrt{D}}$ is a normalization factor to avoid large values after the dot-product operation~\cite{vaswani2017attention}.

To obtain dense correspondences, we use a softmax matching layer~\cite{wang2020learning,kendall2017end,xu2020aanet}, which is not only end-to-end differentiable but also enables sub-pixel accuracy. Specifically, we first normalize the last two dimensions of ${\bm C}_{\mathrm{flow}}$ with the softmax operation, which gives us a distribution
\begin{equation}
    {\bm M}_{\mathrm{flow}} = \mathrm{softmax} ({\bm C}_{\mathrm{flow}}) \in \mathbb{R}^{H \times W \times H \times W}
\end{equation}
for each position in ${\bm F}_1$ with respect to all positions in ${\bm F}_2$. Then, the correspondence $\hat{{\bm G}}_{\mathrm{2D}}$ can be obtained from the weighted average of the matching distribution ${\bm M}_{\mathrm{flow}}$ with the 2D coordinates of pixel grid ${\bm G}_{\mathrm{2D}} \in \mathbb{R}^{H \times W \times 2}$:
\begin{equation}
    \hat{{\bm G}}_{\mathrm{2D}} = {\bm M}_{\mathrm{flow}} {\bm G}_{\mathrm{2D}} \in \mathbb{R}^{H \times W \times 2}.
\end{equation}
Finally, the optical flow ${\bm V}_{\mathrm{flow}}$ can be obtained by computing the difference between the corresponding pixel coordinates:
\begin{equation}
{\bm V}_{\mathrm{flow}} = \hat{{\bm G}}_{\mathrm{2D}}  - {\bm G}_{\mathrm{2D}} \in \mathbb{R}^{H \times W \times 2}.
\end{equation}

\subsubsection{Stereo Matching}
\label{sec:stereo_matching}

Rectified stereo matching aims to find the per-pixel disparity along the horizontal scanline (1D correspondence) between a rectified stereo pair, which can be viewed as a special case of 2D optical flow. Unlike the 2D global matching for optical flow in Eq.~\eqref{eq:flow_corr}, we only need to consider matching along the 1D horizontal direction. More specifically, the correlation matrix for rectified stereo matching is 
\begin{equation}
\label{eq:stereo_corr}
    {\bm C}_{\mathrm{disp}}  \in \mathbb{R}^{H \times W \times W}.
\end{equation}
Similarly, we normalize the last dimension of ${\bm C}_{\mathrm{disp}}$ and obtain the matching distribution
\begin{equation}
    {\bm M}_{\mathrm{disp}} = \mathrm{softmax} ({\bm C}_{\mathrm{disp}}) \in \mathbb{R}^{H \times W \times W}.
\end{equation}
Considering that the correspondence of each pixel in the first image is located to the left of its reference pixel, we mask the upper triangle of the $W \times W$ slices of ${\bm M}_{\mathrm{disp}}$ to avoid unnecessary matches.
Then, the 1D correspondence $\hat{{\bm G}}_{\mathrm{1D}} \in \mathbb{R}^{H \times W}$ can be obtained by computing the weighted average of the matching distribution ${\bm M}_{\mathrm{disp}}$ with all potential horizontal locations ${\bm P} = [0, 1, 2, \cdots, W - 1] \in \mathbb{R}^{W}$:
\begin{equation}
    \hat{{\bm G}}_{\mathrm{1D}} = {\bm M}_{\mathrm{disp}} {\bm P} \in \mathbb{R}^{H \times W}.
\end{equation}
Finally, the (positive) disparity can be obtained by computing the difference between the corresponding coordinates of the 1D horizontal pixel grid ${\bm G}_{\mathrm{1D}} \in \mathbb{R}^{H \times W}$ (which stores only the $x$-coordinates) and $\hat{{\bm G}}_{\mathrm{1D}}$:
\begin{equation}
{\bm V}_{\mathrm{disp}} = {\bm G}_{\mathrm{1D}} - \hat{{\bm G}}_{\mathrm{1D}} \in \mathbb{R}^{H \times W}.
\end{equation}

\subsubsection{Depth Matching}
\label{sec:depth_matching}

For unrectified stereo depth estimation, we assume the camera intrinsic and extrinsic parameters (${\bm K}_1, {\bm E}_1, {\bm K}_2, {\bm E}_2$) for image ${\bm I}_1$ and ${\bm I}_2$ are known (\ie, posed images). They can be obtained via additional sensors like IMU and GPS, or reliably estimated using Structure-from-Motion software like COLMAP~\cite{schonberger2016structure}. To estimate depth, we take an approach similar to the classic plane-sweep stereo method~\cite{collins1996space}. More specifically, we first discretize a predefined depth range $[d_{\mathrm{min}}, d_{\mathrm{max}}]$ as $[d_1, d_2, \cdots, d_N]$ (in our implementation, we discretize the inverse depth domain, while we use depth here for ease of notation). Then for each depth candidate $d_i (i=1, 2, \cdots, N)$, we compute the 2D correspondences $\hat{{\bm G}}_{\mathrm{2D}} \in \mathbb{R}^{H \times W \times 2}$ in ${\bm F}_2$ given the current depth value:
\begin{equation}
\mathcal{H}(\hat{{\bm G}}_{\mathrm{2D}}) = {\bm K}_2 {\bm E}_2 {\bm E}^{-1}_1 d_i {\bm K}^{-1}_1 \mathcal{H}({\bm G}_{\mathrm{2D}}) \in \mathbb{R}^{H \times W \times 3},
\end{equation}
where $\mathcal{H}({\bm G}_{\mathrm{2D}}) \in \mathbb{R}^{H \times W \times 3}$ denotes the homogeneous coordinates of the grid coordinates ${\bm G}_{\mathrm{2D}} \in \mathbb{R}^{H \times W \times 2}$. Next, we perform bilinear sampling on ${\bm F}_2$ with $\hat{{\bm G}}_{\mathrm{2D}}$ and obtain ${\bm F}^i_2 \in \mathbb{R}^{H \times W \times D}$ for depth candidate $d_i$. Their correlation is then computed as
\begin{equation}
    {\bm C}^i = \frac{{\bm F}_1 \cdot {\bm F}^i_2}{\red{\sqrt{D}}}  \in \mathbb{R}^{H \times W}, \quad i = 1, 2, \cdots, N,
\end{equation}
where $\cdot$ is the dot-product operation on the feature dimension $D$. Concatenating the correlations for all depth candidates we obtain
\begin{equation}
    {\bm C}_{\mathrm{depth}} = [{\bm C}^1, {\bm C}^2, \cdots, {\bm C}^N] \in \mathbb{R}^{H \times W \times N}.
\end{equation}
Similar to flow and stereo, we normalize the last dimension of ${\bm C}_{\mathrm{depth}}$ and obtain the matching distribution
\begin{equation}
\label{eq:depth_match}
    {\bm M}_{\mathrm{depth}} = \mathrm{softmax} ({\bm C}_{\mathrm{depth}}) \in \mathbb{R}^{H \times W \times N}.
\end{equation}
Finally, the depth is estimated by computing the weighted average of the matching distribution ${\bm M}_{\mathrm{depth}}$ with all the depth candidates ${\bm G}_{\mathrm{depth}} = [d_1, d_2, \cdots, d_N] \in \mathbb{R}^{N}$:
\begin{equation}
    {\bm V}_{\mathrm{depth}} = {\bm M}_{\mathrm{depth}}{\bm G}_{\mathrm{depth}} \in \mathbb{R}^{H \times W}.
\end{equation}

Thus far, we have presented the detailed matching layers for all three tasks. We remark that all matching layers are differentiable and parameter-free, which not only enables end-to-end training but also doesn't introduce any task-specific learnable parameters. We name our models for flow, stereo and depth tasks GMFlow, GMStereo and GMDepth, respectively, which represent our unified \textit{Global Matching} formulation. Next, we will discuss our model for extracting strong features from the input images.

\subsection{Feature Extraction}
\label{sec:feature}

Key to our formulation lies in obtaining high-quality discriminative features for matching. To achieve this, we combine a common convolutional network (CNN) with a Transformer~\cite{vaswani2017attention} as the feature extractor. More specifically, we first use a weight-sharing ResNet~\cite{he2016deep} to extract $8\times$ downsampled features to keep computation tractable, similar to previous flow methods~\cite{sun2018pwc,teed2020raft}. However, the two features from the CNN are extracted independently, without considering their mutual relations yet. Integrating knowledge from the potential matching candidates in another image can intuitively enhance the feature's distinctiveness and surpass ambiguities, as demonstrated by sparse matching methods~\cite{sarlin2020superglue}. This can be naturally implemented with the cross-attention mechanism, which is able to selectively aggregate information from another image by measuring cross-view feature similarities. We also use a self-attention layer to further improve the feature's quality by considering larger context than the convolutional layer, and a two-layer feed-forward network (FFN, \ie, MLP) to further increase the capacity of the network following the original Transformer~\cite{vaswani2017attention}'s design. The self-attention, cross-attention and FFN constitute a Transformer block, and our final Transformer architecture is a stack of six Transformer blocks which gradually improve the performance (Table.~\ref{tab:flow_conv_vs_softmax}).

Specifically, for the extracted convolutional features $\Tilde{\bm F}_1$ and $\Tilde{\bm F}_2$, we first add fixed 2D sine and cosine positional encodings (following DETR~\cite{carion2020end}) to the features since they lack spatial information. Adding the position information also makes the matching process consider not only the feature similarity but also their spatial distance, which can help resolve ambiguities and improve performance (Table~\ref{tab:transformer}). Then the features are fed into the Transformer for feature enhancement. More specifically, for self-attention, the query, key and value in the attention mechanism~\cite{vaswani2017attention} are the same feature. For cross-attention, the key and value are same but different from the query to model cross-view interactions. This process is performed for both $\Tilde{\bm F}_1$ and $\Tilde{\bm F}_2$ symmetrically:
\begin{equation}
\label{eq:trans_feature}
    {\bm F}_1 = \mathcal{T}(\Tilde{\bm F}_1 + {\bm P}, \Tilde{\bm F}_2 + {\bm P}), \hspace{0.6em} {\bm F}_2 = \mathcal{T}(\Tilde{\bm F}_2 + {\bm P}, \Tilde{\bm F}_1 + {\bm P}),
\end{equation}
where $\mathcal{T}$ is a Transformer, ${\bm P}$ is the positional encoding, the first input of $\mathcal{T}$ is query and the second is key and value.

One issue with the standard Transformer architecture~\cite{vaswani2017attention} is the quadratic computational complexity due to the pair-wise attention operation. To improve efficiency, we adopt the shifted local window attention strategy from Swin Transformer~\cite{liu2021Swin}. However, unlike Swin that uses a \emph{fixed window size}, we split the feature to \emph{fixed number of local windows} to make the window size adaptive with the feature's spatial size. In this way, the attention mechanism can model long-range dependencies on high-resolution feature maps and accordingly better performance for large displacements can be achieved.  Specifically, we split the input feature of size $H \times W$ to $K \times K$ windows (each with size $\frac{H}{K} \times \frac{W}{K}$, better for large displacements flow if $K$ is smaller, see Table~\ref{tab:split_attn}), and perform self- and cross-attentions within each local window independently. For every two consecutive local windows, we shift the window partition by $(\frac{H}{2K}, \frac{W}{2K})$ to introduce cross-window connections. In our method, we split into $2 \times 2$ windows (each with size $\frac{H}{2} \times \frac{W}{2}$), which leads to a good speed-accuracy trade-off (Table~\ref{tab:split_attn}).

\begin{figure}[t]
    \centering
    \includegraphics[width=0.98\linewidth]{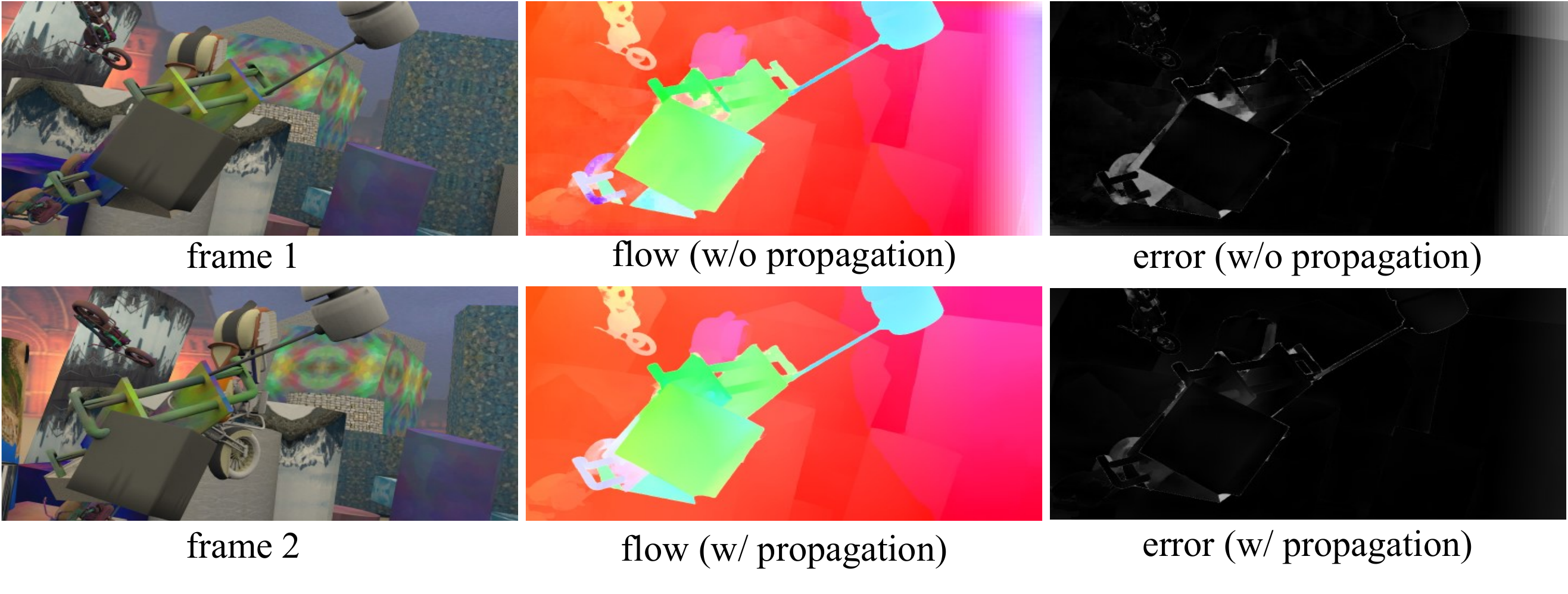}
    \caption{\textbf{Our propagation strategy significantly improves the performance of occluded and out-of-boundary pixels.}}
    \label{fig:flow_prop}
    \vspace{-10pt}
\end{figure}

We note that for the rectified stereo matching task, since the correspondences lie only on the 1D horizontal direction, it's redundant to perform full 2D cross-attention in the Transformer. Thus we perform 1D horizontal cross-attention for the stereo matching task, which is not only faster, but also leads to better performance (Table~\ref{tab:stereo_1d_2d_attn}). Since only the linear feature projection layers of the Transformer are learnable, the final models are not affected by the specific parameter-free attention operation (2D, 1D or any other forms). Thus all the learnable parameters remain exactly the same for all three tasks.

\subsection{Propagation}
\label{sec:prop}

Our matching-based formulation implicitly assumes that corresponding pixels are visible in both images and thus they can be matched by comparing their similarities. However, this assumption will be invalid for occluded and out-of-boundary pixels, producing unreliable results in these regions (Fig.~\ref{fig:flow_prop}). To remedy this, by observing that the flow/disparity/depth field and the image itself share high structure similarity~\cite{hui2020liteflownet3,Jiang_2021_ICCV}, we propose to propagate the high-quality flow/disparity/depth predictions to unmatched regions by measuring feature self-similarity. This operation can be implemented efficiently with a single self-attention layer (illustrated in Fig.~\ref{fig:overview}):
\begin{equation}
    \hat{\bm V} = \mathrm{softmax} \left(\frac{{\bm F}_1 {\bm F_1}^T}{\sqrt{D}}\right) {\bm V}  \in \mathbb{R}^{H \times W \times 2},
\end{equation}
where ${\bm V}$ is the flow/disparity/depth prediction from the softmax matching layer in Sec.~\ref{sec:formulation}. Note that we don't explicitly differentiate matched and unmatched pixels, but simply learn such a propagation process with ground truth flow supervision. Fig.~\ref{fig:flow_prop} shows that this strategy can effectively correct the errors in unmatched regions.

Our current estimate is at the $8 \times$ downsampled feature resolution. To get the original image resolution prediction, we use RAFT's upsampling~\cite{teed2020raft} method that computes the full resolution flow/disparity/depth at each pixel as a weighted combination of a $3 \times 3$ grid of its coarse resolution neighbors. The combination weights are learned with a small 2-layer convolutional network, whose output channel is $8\times 8 \times 3 \times 3$ for $8\times$ upsampling. Fig.~\ref{fig:overview} provides an overview of our unified model.

\subsection{Refinement}

Our method presented so far (based on $1/8$ features) already achieves competitive performance while being simple and efficient. It can be further improved by using additional refinement steps, yielding different speed-accuracy trade-offs. We explore two types of refinement in this paper: hierarchical matching refinement with higher-resolution ($1/4$) features and local regression refinement with convolutions. {We remark that the hierarchical matching refinement uses our matching-based formulation and thus is task-agnostic, while the local regression refinement is task-specific but optional. It can hence be viewed as a post-processing step to further improve the performance of our unified method.}

\subsubsection{Hierarchical Matching Refinement}
\label{sec:hierarchical_match}

Our unified global matching is performed at $1/8$ feature resolution, and a $1/8$ flow/disparity/depth prediction is obtained. Using additional higher-resolution ($1/4$) features for matching can further improve the performance and fine-grained details, while not introducing any task-specific learnable parameters as it uses our matching-based formulation. However, we found the improvement for unrectified stereo depth estimation to be not as significant as flow and stereo, and thus we choose to not perform hierarchical matching at $1/4$ resolution for the depth task. Specifically, for optical flow and rectified stereo matching tasks, we first upsample the $1/8$ flow/disparity prediction to $1/4$ resolution, and then warp the second CNN feature with the upsampled flow/disparity. In this way, the remaining task is reduced to matching between the original first CNN feature and the warped second CNN feature, and thus the same model depicted in Fig.~\ref{fig:overview} can be used at $1/4$ resolution but in a local range for refinement. More specifically, we perform a $9 \times 9$ local window matching for optical flow, and 1D horizontal local matching with length $9$ for stereo matching (similar formulations as Sec.~\ref{sec:flow_matching} and Sec.~\ref{sec:stereo_matching} but in a local range). The predicted flow/disparity residual is then added to the previous upsampled flow/disparity prediction obtained by global matching. For the Transformer, we split the $1/4$ feature map to $8 \times 8$ local windows (each with $1/32$ of the original image resolution) in attention computation to model local-range interactions. Next, we perform a $3 \times 3$ local window self-attention operation for flow/disparity propagation (similar formulation as Sec.~\ref{sec:prop} but in a local range). Finally, the $1/4$ flow/disparity prediction is obtained and it's upsampled to the full resolution.

We note that we share the Transformer and self-attention weights in the $1/8$ and $1/4$ hierarchical matching stages since they perform basically very similar matching process except for different ranges (global \vs \ local). This not only reduces parameters but also improves generalization, as shown in the original GMFlow~\cite{xu2022gmflow} paper. To generate the $1/4$ and $1/8$ resolution features, we take a similar approach to TridentNet~\cite{li2019scale}. Specifically, we first obtain a $1/4$ resolution feature map with a CNN, and then append a weight-sharing $3 \times 3$ convolution with strides $1$ and $2$ to generate two-branch features at $1/4$ and $1/8$ resolutions, respectively. Such a weight-sharing design also leads to less parameters and better performance than using feature pyramid network~\cite{lin2017feature} (see GMFlow~\cite{xu2022gmflow} paper)

\begin{table*}[t]
    \centering
    \setlength{\tabcolsep}{2.5pt} %
    \begin{tabular}{lcccccccccccccccccc}
    \toprule
    \multirow{2}{*}[-2pt]{Method} & \multirow{2}{*}[-2pt]{\#blocks} & \multicolumn{4}{c}{Things (val, clean)} & \multicolumn{4}{c}{Sintel (train, clean)} & \multicolumn{4}{c}{Sintel (train, final)} &  \multirow{2}{*}[-2pt]{\begin{tabular}[x]{@{}c@{}}Param\\(M) \end{tabular}} \\
    \addlinespace[-10pt] \\
    \cmidrule(lr){3-6} \cmidrule(lr){7-10} \cmidrule(lr){11-14}
    \addlinespace[-10pt] \\
    & & EPE & $s_{0-10}$ & $s_{10-40}$ & $s_{40+}$ & EPE & $s_{0-10}$ & $s_{10-40}$ & $s_{40+}$ & EPE & $s_{0-10}$ & $s_{10-40}$ & $s_{40+}$ & \\
    
    \midrule

    \multirow{4}{*}[-2pt]{cost volume + conv} & 0 & 18.83 & 3.42 & 6.49 & 49.65 & 6.45 & 1.75 & 7.17 & 38.19 & 7.75 & 2.10 & 8.88 & 45.29 & 1.8 \\
    & 4 & 10.99 & 1.70 & 3.41 & 29.78 & 3.32 & 0.73 & 3.84 & 20.58 & 4.93 & 0.99 & 5.71 & 31.16 & 4.6 \\
    & 8 & 9.59 & 1.44 & 2.96 & 26.04 & 2.89 & 0.65 & 3.36 & 17.75 & 4.32 & 0.88 & 4.95 & 27.33 & 8.0 \\
    & 12 & 9.04 & 1.37 & 2.84 & 24.46 & 2.78 & 0.65 & 3.32 & 16.69 & 4.07 & 0.84 & 4.76 & 25.44 & 11.5 \\
    & 18 & 8.67 & 1.33 & 2.74 & 23.43 & 2.61 & 0.59 & 3.07 & 15.91 & 3.94 & 0.82 & 4.62 & 24.58 & 15.7 \\

    \midrule
    
    \multirow{4}{*}[-2pt]{Transformer + softmax} & 0 & 22.93 & 8.57 & 11.13 & 52.07 & 8.44 & 2.71 & 11.60 & 42.10 & 10.28 & 3.11 & 13.83 & 53.34 & 1.0 \\
    & 1 & 11.45 & 2.98 & 4.68 & 28.35 & 4.12 & 1.27 & 5.08 & 22.25 & 6.11 & 1.70 & 7.89 & 33.52 & 1.6 \\
    & 2 & 8.59 & 1.80 & 3.28 & 21.99 & 3.09 & 0.90 & 3.66 & 17.37 & 4.54 & 1.24 & 5.44 & 26.00 &  2.1 \\  %
    & 4 & 7.19 & 1.40 & 2.62 & 18.66 & 2.43 & 0.67 & 2.73 & 14.23 & 3.78 & 1.01 & 4.27 & 22.37 & 3.1 \\  %
    & 6 & \textbf{6.67} & \textbf{1.26} & \textbf{2.40} & \textbf{17.37} & \textbf{2.28} & \textbf{0.58} & \textbf{2.49} & \textbf{13.89} & \textbf{3.44} & \textbf{0.80} & \textbf{3.97} & \textbf{21.02} & 4.2 \\  %
    
    \midrule

    conv + softmax & 6 & 17.06 & 5.79 & 7.74 & 40.03 & 6.36 & 2.15 & 8.53 & 31.53 & 8.00 & 2.45 & 10.42 & 42.09 & 5.1\\
    
    \bottomrule
    \end{tabular}
    \caption{\textbf{Methodology comparison for optical flow task}. We stack different numbers of convolutional residual blocks or Transformer blocks to see how performance varies. All models are trained on Chairs and Things training sets. We report the performance on Things (clean) validation set and cross-dataset generalization results on Sintel (clean and final) training sets. Our method outperforms previous cost volume and convolution-based approach by a large margin, especially for large motions ($s_{40+}$). Replacing the Transformer in our model with a convolutional network (\ie, conv + softmax) leads to a significant performance drop, since convolutions are not able to model cross-view interactions (which is important for obtaining high-quality discriminative features, see also Table~\ref{tab:transformer}). 
    }
    \label{tab:flow_conv_vs_softmax}
\end{table*}

\subsubsection{Local Regression Refinement}
\label{sec:local_reg_refine}

Our unified model thus far with only one task-agnostic hierarchical matching refinement is able to outperform RAFT~\cite{teed2020raft} with 31 refinements while running faster (Fig.~\ref{fig:flow_iter_vs_epe} and Table~\ref{tab:flow_raft_vs_gmflow}), which demonstrates the effectiveness and efficiency of our global matching-based formulation. However, our matching-based method is also complementary to previous cost volume and convolution-based local regression approach. We observe the strength of our unified matching method mainly in the presence of large motion (Table~\ref{tab:flow_conv_vs_softmax} and \ref{tab:flow_raft_vs_gmflow}). For small motions, it might not be necessary to perform global matching (Table~\ref{tab:global_local_match}) and in this case, local regression is advantageous. To achieve the best system-level performance, one straightforward way is to combine the strengths of these two kinds of flow estimation approaches. That is, the local regression method is used as a post-processing step to our unified model. This further improves fine-grained details and regions that are hard to match.

More specifically, we use RAFT's iterative refinements for further improvement. At each refinement, we produce an update based on the current prediction. The update is regressed with convolutions from local correlations. The local correlations are task-specific: for optical flow, we use 2D correlation; for rectified stereo matching, we also use 2D correlation since we found it to perform better than 1D correlation (Table~\ref{tab:stereo_1d_2d_attn}) although some redundancy exists; for unrectified stereo depth, we use 2D correlation constructed from the current depth prediction and relative camera transformation. Such refinement architectures are task-specific and not shared across tasks. The optional number of additional iterative refinements is also different for different tasks, we choose this number empirically. More specifically, for optical flow, we use 6 additional refinement steps at $1/4$ feature resolution after the hierarchical matching refinement; for rectified stereo matching, we use 3 additional refinement steps at $1/4$ feature resolution after the hierarchical matching refinement; for unrectified stereo depth estimation, we use 1 additional refinement step at $1/8$ feature resolution and no hierarchical matching is used. Note that the number of refinement steps as part of our post-processing is much less than previous pure iterative architectures (\eg, 31 refinements in RAFT~\cite{teed2020raft} and its recent variants~\cite{Jiang_2021_ICCV,sui2022craft,huang2022flowformer}) thanks to our stronger base model.

\subsection{Training Loss}

We supervise all predictions (including the intermediate network outputs and final ones) with the ground truth:
\begin{equation}
    L = \sum_{i=1}^{N} \gamma^{N - i} \ell ({\bm V}_i, {\bm V}_{\mathrm{gt}}),
\end{equation}
where $N$ is the total number of predictions, and $\gamma$ (set to 0.9) is the weight that is exponentially increasing to give higher weights for later predictions following RAFT~\cite{teed2020raft}.

The definition of the loss function $\ell$ are following previous methods. More specifically, for optical flow, we use an $L_1$ loss~\cite{teed2020raft}; for rectified stereo matching, we use the smooth $L_1$ loss~\cite{xu2020aanet}; for unrectified stereo depth estimation, we use the $L_1$ loss on the inverse depth~\cite{ummenhofer2017demon}. Following~\cite{Teed2020DeepV2D}, we also use an additional gradient loss for unrectified stereo depth:
\begin{equation}
    L_{\mathrm{grad}} = \sum_{i=1}^{N} \gamma^{N - i} (\ell (\partial_x {\bm V}_i, \partial_x {\bm V}_{\mathrm{gt}}) + \ell (\partial_y {\bm V}_i, \partial_y {\bm V}_{\mathrm{gt}})),
\end{equation}
where $\ell$ is the $L_1$ loss. The total loss for the depth task is a combination of the inverse depth loss and the gradient loss, where the combination weights are both 20.

\section{Experiments}

In this section, we will first study the properties of our unified model for each task independently, and then show the unique advantage of our unified model by cross-task transfer, and finally perform system-level comparisons with previous methods on standard benchmarks. 

\noindent \textbf{Implementation details.} We implement our full model in PyTorch. Our convolutional backbone network is a ResNet-like~\cite{he2016deep} architecture and the output feature is $1/8$ of the original image resolution for the default model. When hierarchical matching is used, the output features are of $1/4$ and $1/8$ resolutions as mentioned in Sec.~\ref{sec:hierarchical_match}. The feature dimension is $128$. We stack 6 Transformer blocks, where each Transformer block consists of a self-attention layer, a cross-attention layer and a feed-forward network. We only use a single head in all the attention computations, since we didn't observe obvious performance gains with multi-head attention. We use the AdamW~\cite{loshchilov2017decoupled} optimizer and a cosine learning rate scheduler with warmup to optimize the model.

\subsection{Optical Flow}

\noindent \textbf{Datasets and evaluation setup.} Following previous optical flow methods~\cite{ilg2017flownet,sun2018pwc,teed2020raft}, we first train on the FlyingChairs (Chairs)~\cite{dosovitskiy2015flownet} and FlyingThings3D (Things)~\cite{mayer2016large} datasets, and then evaluate on Sintel~\cite{butler2012naturalistic} and KITTI~\cite{menze2015object} training sets for cross-dataset generalization. We also evaluate on the Things validation set to see how the model performs on the same-domain data. Finally, we perform additional fine-tuning on Sintel and KITTI training sets and report the performance on the online benchmarks.

\noindent \textbf{Metrics.} We adopt the commonly used metric in optical flow, \ie, the end-point-error (EPE), which is the average $\ell_2$ distance between the prediction and ground truth. For the KITTI dataset, we also use \emph{F1-all}, which reflects the percentage of outliers. To better understand the performance gains, we also report the EPE for different motion magnitudes. 
Specifically, we use $s_{0-10}$, $s_{10-40}$ and $s_{40+}$ to denote the EPE over pixels with ground truth flow motion magnitude falling into the ranges of $0-10$, $10-40$ and more than $40$ pixels, respectively.

\noindent \textbf{Training schedule.} For methodology comparison (Sec.~\ref{sec:method_comp}) and ablation experiments (Sec.~\ref{sec:ablation}), we first train our default $1/8$ feature resolution model on Chairs dataset for 100K iterations, with a batch size of 16 and a learning rate of 4e-4. We then finetune the model on Things dataset for 200K iterations, with a batch size of 8 and a learning rate of 2e-4. The models thus far all use bilinear upsamling to upsample to the full resolution flow prediction for simplicity. For later experiments, we use RAFT's convex upsampling~\cite{teed2020raft} and our models are trained on Things dataset for 800K iterations, which leads to improved performance~\cite{xu2022gmflow}. For the final fine-tuning process on Sintel and KITTI datasets for benchmark comparisons, we report details in Sec.~\ref{sec:flow_benchmark}.

\begin{table*}[t]
\centering
\subfloat[
\textbf{Transformer components}. Cross-attention contributes most.
\label{tab:transformer}
]{
\centering
\begin{minipage}{0.45\linewidth}
{\begin{center}
\begin{tabular}{lccccccccccccccc}
    \toprule
    
    \multirow{2}{*}[-2pt]{setup} & \multicolumn{1}{c}{Things (val)} & \multicolumn{2}{c}{Sintel (train)} &  \multirow{2}{*}[-2pt]{\begin{tabular}[x]{@{}c@{}}Param\\(M) \end{tabular}} \\
    \addlinespace[-10pt] \\
    \cmidrule(lr){2-2} \cmidrule(lr){3-4}
    \addlinespace[-10pt] \\
    & clean & clean & final & \\
    \midrule

    full & \textbf{6.67} & \textbf{2.28} & \textbf{3.44} & 4.2 \\
    
    w/o cross attn. & 10.84 & 4.48 & 6.32 & 3.8  \\
    w/o position & 8.38 & 2.85 & 4.28 & 4.2 \\
    w/o FFN & 8.71 & 3.10 & 4.43 & 1.8 & \\
    w/o self attn. & 7.04 & 2.49 & 3.69 & 3.8 \\
    
    \bottomrule
    \end{tabular}
\end{center}}
\end{minipage}
}
\hspace{2em}
\subfloat[
\textbf{Numbers of window splits in shifted local attention}. $2 \times 2$ represents a good speed-accuracy trade-off.
\label{tab:split_attn}
]{
\begin{minipage}{0.45\linewidth}{\begin{center}
\begin{tabular}{cccccccccccccccc}
    \toprule
    
    \multirow{2}{*}[-2pt]{\#splits} & \multicolumn{4}{c}{Things (val, clean)}  &  \multirow{2}{*}[-2pt]{\begin{tabular}[x]{@{}c@{}}Time\\(ms) \end{tabular}} \\
    \addlinespace[-10pt] \\
    \cmidrule(lr){2-5} 
    \addlinespace[-10pt] \\
    & EPE & $s_{0-10}$ & $s_{10-40}$ & $s_{40+}$  \\
    \midrule
    
    $1 \times 1$ & 6.34 & 1.26 & 2.37 & 16.36 & 105  \\
    \underline{$2 \times 2$} & 6.67 & 1.26 & 2.40 & 17.37 & 53 \\
    $4 \times 4$ & 7.32 & 1.29 & 2.58 & 19.26 & 35 \\
    \bottomrule
    \\

    \end{tabular}
\end{center}}\end{minipage}
}
\\
\subfloat[
\textbf{Global \vs \ local matching}. Global matching is significantly better for large motions while being fast to compute.
\label{tab:global_local_match}
]{
\begin{minipage}{0.45\linewidth}{\begin{center}
\setlength{\tabcolsep}{3pt} %
\begin{tabular}{ccccccccccccccc}
    \toprule
    
    \multirow{2}{*}[-2pt]{\begin{tabular}[x]{@{}c@{}}matching \\space \end{tabular}} & \multicolumn{4}{c}{Things (val, clean)}  & \multirow{2}{*}[-2pt]{\begin{tabular}[x]{@{}c@{}}Time\\(ms) \end{tabular}} \\
    \addlinespace[-10pt] \\
    \cmidrule(lr){2-5} 
    \addlinespace[-10pt] \\
    & EPE & $s_{0-10}$ & $s_{10-40}$ & $s_{40+}$  \\
    \midrule
    
    global & \textbf{6.67} & 1.26 & \textbf{2.40} & \textbf{17.37} & 52.6 \\
    local $3 \times 3$ & 31.78 & 1.19 & 12.40 & 85.39 & 51.2 \\
    local $5 \times 5$ & 26.51 & \textbf{0.89} & 6.67 & 76.76 & 51.5 \\
    local $ 9 \times 9$ & 19.88 & 1.01 & 2.44 & 61.06 & 52.9 \\

    \bottomrule
    \end{tabular}
\end{center}}\end{minipage}
}
\hspace{2em}
\subfloat[
\textbf{Flow propagation} greatly improves unmatched pixels. 
\label{tab:prop}
]{
\begin{minipage}{0.45\linewidth}{\begin{center}
\setlength{\tabcolsep}{1pt} %
\begin{tabular}{lccccccccccccccc}
    \toprule
    
    \multirow{2}{*}[-2pt]{prop.} & \multicolumn{3}{c}{Sintel (clean)}  &  \multicolumn{3}{c}{Sintel (final)} \\
    \addlinespace[-10pt] \\
    \cmidrule(lr){2-4} \cmidrule(lr){5-7} 
    \addlinespace[-10pt] \\
    & all & matched & unmatched & all & matched & unmatched  \\
    \midrule
    
    w/o  & 2.28 & \textbf{1.06} & 15.54 & 3.44 & \textbf{1.95} & 19.50  \\
    w/  & \textbf{1.89} & 1.10 & \textbf{10.39} & \textbf{3.13} & 1.98 & \textbf{15.52} \\

    \bottomrule
    \\

    \end{tabular}
\end{center}}\end{minipage}
}
\vspace{-10pt}
\caption{\textbf{GMFlow ablations for optical flow task}. All models are trained on Chairs and Things training sets.}
\label{tab:flow_ablations}
\end{table*}

\subsubsection{Methodology Comparison}
\label{sec:method_comp}

\noindent {\bf Flow estimation approach.} We compare our Transformer and softmax-based flow estimation method with cost volume and convolution-based approaches. Specifically, we adopt the state-of-the-art cost volume construction method in RAFT~\cite{teed2020raft} that concatenates 4 local cost volumes at 4 scales, where each cost volume has a dimension of $H \times W \times (2R+1)^2$. Here $H$ and $W$ denote the feature's spatial size, and the search range $R$ is set to 4 following RAFT. To regress flow, we stack different numbers of convolutional residual blocks~\cite{he2016deep} to see how the performance varies. The final optical flow is obtained with a $3 \times 3$ convolution with 2 output channels. For our method, we stack different numbers of Transformer blocks for feature enhancement and the final optical flow is obtained with a global correlation and softmax layer. Table~\ref{tab:flow_conv_vs_softmax} shows that the performance improvement of our method is more significant compared to the cost volume and convolution-based approach. For instance, our method with 2 Transformer blocks is already able to outperform 8 convolution blocks, especially in the presence of large motions ($s_{40+}$). The performance can be further improved by stacking more layers, surpassing the cost volume and convolution-based approach by a large margin. We also replace the Transformer in our model with a convolutional network for feature enhancement, which leads to a large drop in performance. This is largely due to the unique advantage of the cross-attention mechanism for modeling cross-view interactions (see Table~\ref{tab:transformer} for detailed evaluations of the Transformer components), which enables aggregation the information from the other frame by considering cross-view similarities and thus greatly improves the quality of the extracted features. This is not achievable with convolutions~\cite{Zbontar2016Stereo}.

\begin{figure}[t]
    \centering
    \includegraphics[width=0.98\linewidth]{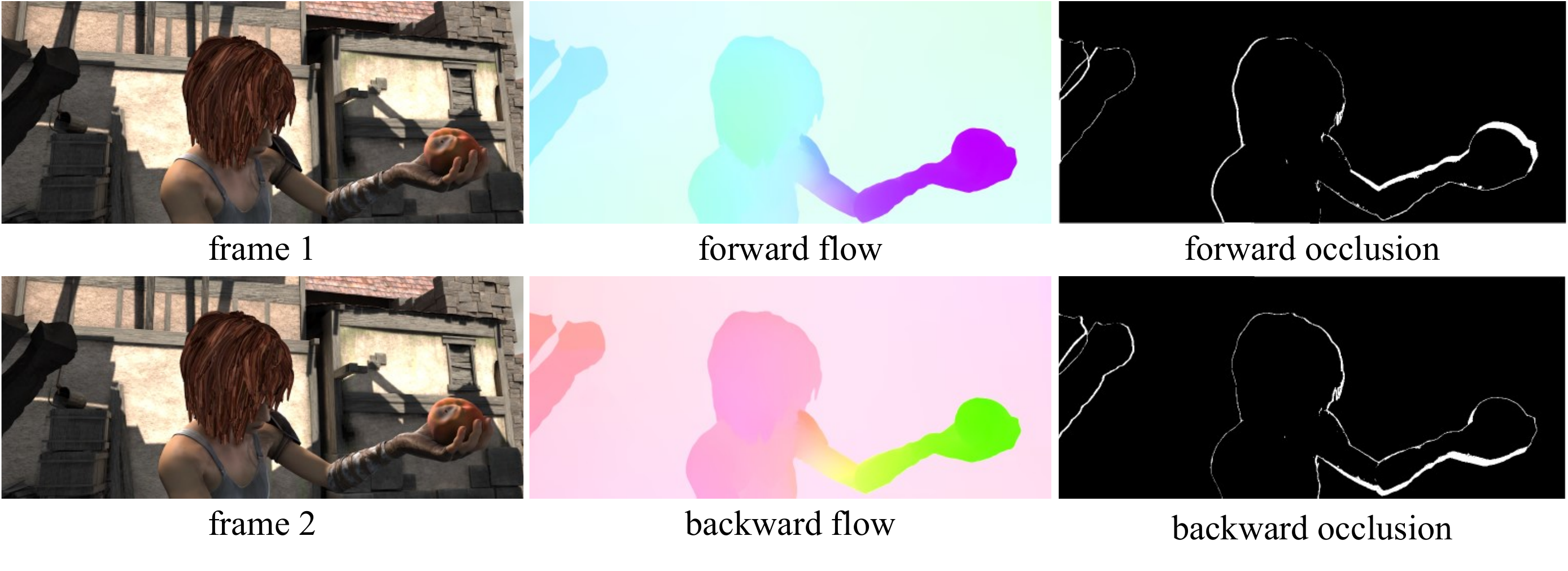}
    \caption{\textbf{GMFlow simplifies backward flow computation} by directly transposing the global correlation matrix without requiring to forward the network twice. The bidirectional flow can be used for occlusion detection with forward-backward consistency check.}
    \label{fig:bidir_flow}
    \vspace{-10pt}
\end{figure}

\noindent {\bf Bidirectional flow prediction.} Our method also simplifies backward optical flow computation by directly transposing the global correlation matrix in Eq.~\eqref{eq:flow_corr}. Note that during training we only predict unidirectional flow while at inference, we can obtain bidirectional flow for free, without requiring to forward the network twice, unlike previous regression-based methods~\cite{meister2018unflow,hur2019iterative}. The bidirectional flow can be used for occlusion detection with forward-backward consistency check (following~\cite{meister2018unflow}), as shown in Fig.~\ref{fig:bidir_flow}.

\begin{table*}[t]
    \centering
    \setlength{\tabcolsep}{3.pt} %
    \begin{tabular}{lccccccccccccccr}
    \toprule
    \multirow{2}{*}[-2pt]{Method} & \multirow{2}{*}[-2pt]{\#refine.} & \multicolumn{4}{c}{Things (val, clean)} & \multicolumn{4}{c}{Sintel (train, clean)} & \multicolumn{4}{c}{Sintel (train, final)} & \multirow{2}{*}[-2pt]{\begin{tabular}[x]{@{}c@{}}Param\\(M) \end{tabular}}   & \multirow{2}{*}[-2pt]{\begin{tabular}[x]{@{}c@{}}Time\\(ms) \end{tabular}}  \\
    \addlinespace[-10pt] \\
    \cmidrule(lr){3-6} \cmidrule(lr){7-10} \cmidrule(lr){11-14}
    \addlinespace[-10pt] \\
    & & EPE & $s_{0-10}$ & $s_{10-40}$ & $s_{40+}$ & EPE & $s_{0-10}$ & $s_{10-40}$ & $s_{40+}$ & EPE & $s_{0-10}$ & $s_{10-40}$ & $s_{40+}$ & \\
    
    \midrule
    
    \multirow{6}{*}[-2pt]{RAFT~\cite{teed2020raft}} & 0 & 14.28 & 1.47 & 3.62 & 40.48 & 4.04 & 0.77 & 4.30 & 26.66 & 5.45 & 0.99 & 6.30 & 35.19 & \multirow{6}{*}[-2pt]{5.3} & 25 (14) \\
    & 3 & 6.27 & 0.69 & 1.67 & 17.63 & 1.92 & 0.47 & 2.32 & 11.37 & 3.25 & 0.65 & 4.00 & 20.04 & & 39 (21) \\
    & 7 & 4.66 & 0.55 & 1.38 & 12.87 & 1.61 & 0.39 & 1.90 & 9.61 & 2.80 & 0.53 & 3.30 & 17.76 & & 58 (31) \\
    & 11 & 4.31 & 0.53 & 1.33 & 11.79 & 1.55 & 0.41 & 1.73 & 9.19 & 2.72 & 0.52 & 3.12 & 17.43 & & 78 (41) \\
    & 23 & 4.22 & 0.53 & 1.32 & 11.52 & 1.47 & 0.36 & 1.63 & 9.00 & 2.69 & 0.52 & 3.05 & 17.28 & & 133 (71)\\
    & 31 & 4.25 & \textbf{0.53} & 1.31 & 11.63 & 1.41 & 0.32 & 1.55 & 8.83 & 2.69 & 0.52 & 3.00 & 17.45 & & 170 (91) \\
    
    \midrule

     \multirow{2}{*}[-2pt]{{GMFlow}} & 0 & 3.48 & 0.67 & 1.31  & 8.97 & 1.50 & 0.46  & 1.77 & 8.26  & 2.96  & 0.72 & 3.45  & 17.70 & 4.7 & 57 (26) \\
     & 1 & \textbf{2.80} & \textbf{0.53} & \textbf{1.01} & \textbf{7.31} & \textbf{1.08} & \textbf{0.30} & \textbf{1.25} & \textbf{6.26} & \textbf{2.48} & \textbf{0.51}  & \textbf{2.81} & \textbf{15.67} & 4.7 & 151 (66) \\

    \bottomrule
    \end{tabular}
    \caption{\textbf{RAFT's iterative refinement architecture \textit{vs.} our GMFlow model}. The models are trained on Chairs and Things training sets. The inference time is measured on a single V100 and A100 (in parentheses) GPU at Sintel resolution ($436\times 1024$). Our method gains more speedup than RAFT ($2.29\times$ \textit{vs.} $1.87 \times$, \ie, ours: $151 \to 66$, RAFT: $170 \to 91$) on the high-end A100 GPU since our method doesn't require a large number of sequential computation.
    }
    \label{tab:flow_raft_vs_gmflow}
    \vspace{-10pt}
\end{table*}

\subsubsection{Ablations}
\label{sec:ablation}

\noindent {\bf Transformer components.} We ablate different Transformer components in Table~\ref{tab:transformer}. The cross-attention contributes most, since it models the cross-view interactions between two features, which integrates the knowledge from another image and greatly improves the quality of the extracted features. Also, the position information makes the matching process position-dependent, which can help alleviate the ambiguities in pure feature similarity-based matching. Removing the feed-forward network (FFN) reduces a large number of parameters, while also leading to a moderate performance drop. The self-attention aggregates contextual cues within the same feature, leading to additional gains.

\noindent {\bf Local window attention.} We compare the speed-accuracy trade-off of splitting the features into different numbers of local windows for attention computation in Table~\ref{tab:split_attn}. Recall that the extracted features from our CNN backbone have a resolution of $1/8$, further splitting into $H/2 \times W /2$ local windows (\ie, $1/16$ of the original image resolution) leads to a good trade-off between accuracy and speed, and thus is used in our model.

\noindent {\bf Matching Space.} We replace our global matching (\ie, all pair-wise matching $H \times W \times H \times W$ in Eq.~\eqref{eq:flow_corr}) with local matching (\ie, reduce the global matching in Eq.~\eqref{eq:flow_corr} to a local one $H \times W \times K \times K$ with window size $K \times K$) in Table~\ref{tab:global_local_match} and observe a significant performance drop, especially for large motion ($s_{40+}$). Besides, global matching can be computed efficiently with a simple matrix multiplication, while a larger size for local matching will be slower due to the excessive sampling operation.

\noindent {\bf Flow propagation.} Our flow propagation strategy results in significant performance gains in unmatched regions (including occluded and out-of-boundary pixels), as shown in Table~\ref{tab:prop} and Fig.~\ref{fig:flow_prop}. The structural correlation between the feature and flow provides a valuable cue to improve the performance of pixels that are challenging to match.

\begin{figure}[t]
    \centering
    \includegraphics[width=0.9\linewidth]{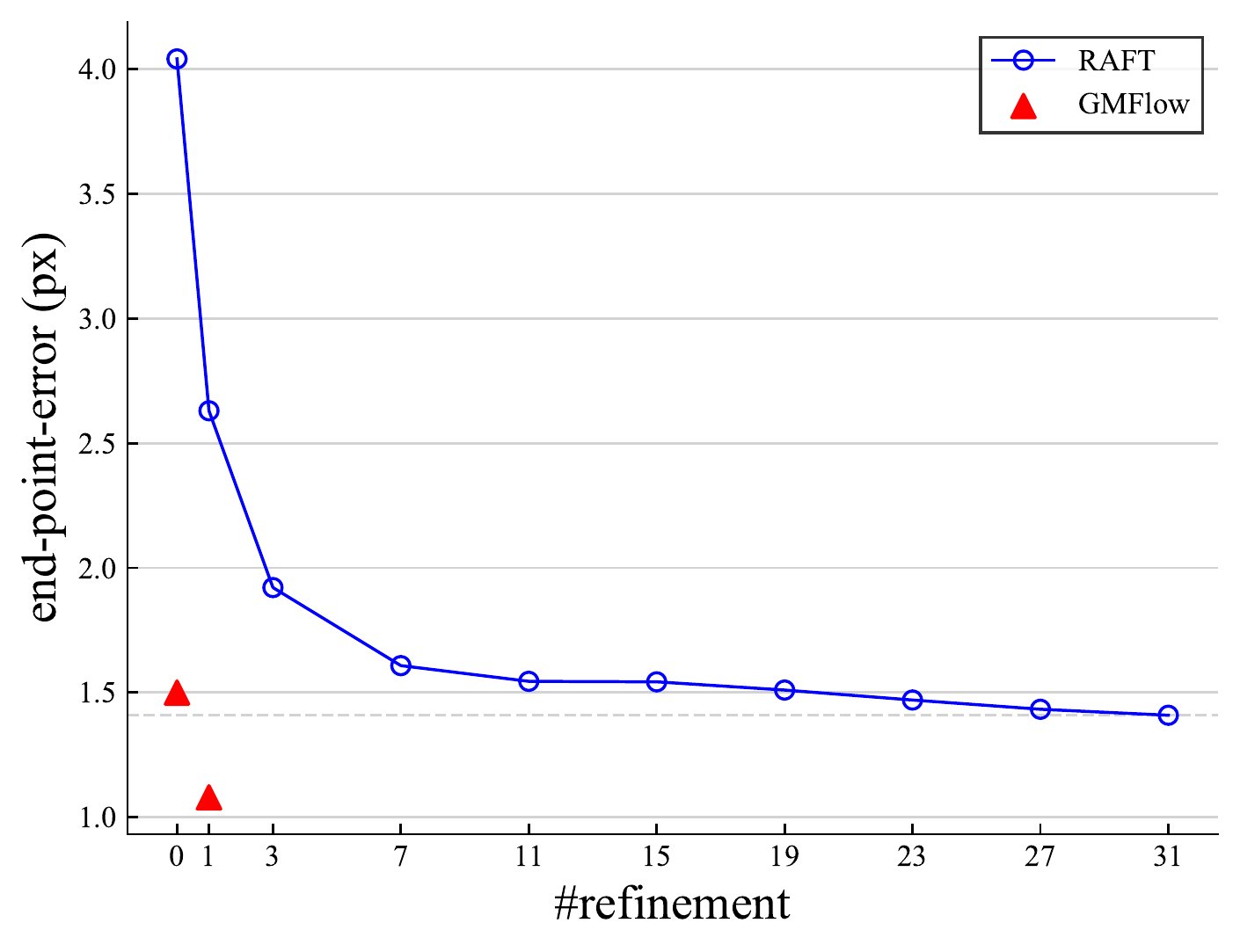}
    \caption{Optical flow \textbf{end-point-error \vs \ number of refinements} at inference time. This figure shows the generalization on Sintel (clean) training set after training on Chairs and Things datasets. 
    Our unified model with only one hierarchical matching refinement outperforms RAFT with 31 refinement steps while running faster (see Table~\ref{tab:flow_raft_vs_gmflow}).
    }
    \label{fig:flow_iter_vs_epe}
    \vspace{-10pt}
\end{figure}

\begin{table*}[t]
\centering
\subfloat[
\textbf{Rectified stereo matching task}.
\label{tab:stereo_ablation}
]{
\centering
\begin{minipage}{0.43\linewidth}
{\begin{center}
\begin{tabular}{lccccccccccccc}
    \toprule

    \multirow{2}{*}[-2pt]{setup} &  \multicolumn{2}{c}{Things} &
    \multicolumn{2}{c}{KITTI} & \multirow{2}{*}[-2pt]{\begin{tabular}[x]{@{}c@{}}Param\\(M) \end{tabular}} \\
    \addlinespace[-10pt]  \\
    \cmidrule(lr){2-3} \cmidrule(lr){4-5} 
    \addlinespace[-10pt] \\
    & EPE & D1 & EPE & D1 & \\
    
    \midrule
    
    full & \textbf{1.22} & \textbf{3.70} & \textbf{1.61} & 10.53 & 4.7 \\
    w/o cross attn. & 1.96 & 7.84 & 5.40 & 31.97 & 4.3 \\
    w/o position & 1.24 & 4.46 & 1.72 & 12.07 & 4.7 \\
    w/o FFN & 1.39 & 4.93 & 1.95 & 14.86 & 2.3 \\
    w/o self attn. & 1.35 & 4.47 & 1.87 & 13.04 & 4.3 \\
    w/o propagation & 2.33 & 6.08 & 1.76 & \textbf{10.09} & 4.6 \\

    \bottomrule
    \end{tabular}
\end{center}}
\end{minipage}
}
\hspace{2em}
\subfloat[
\textbf{Unrectified stereo depth estimation task}.
\label{tab:depth_ablation}
]{
\begin{minipage}{0.5\linewidth}{
\begin{center}
\setlength{\tabcolsep}{2pt} %
\begin{tabular}{lccccccccccccc}
    \toprule

    setup & Abs Rel & Sq Rel & RMSE & RMSE log & Param (M) \\
    
    \midrule
    
    full & \textbf{0.074} & \textbf{0.028} & \textbf{0.225} & \textbf{0.103} & 4.7 \\
    w/o cross attn. & 0.095 & 0.043 & 0.284 & 0.132 & 4.3 \\
    w/o position & 0.078 & 0.031 & 0.237 & 0.109 & 4.7 \\
    w/o FFN & 0.089 & 0.041 & 0.276 & 0.127 & 2.3 \\
    w/o self attn. & 0.081 & 0.034 & 0.248 & 0.114 & 4.3 \\
    w/o propagation & 0.091 & 0.045 & 0.293 & 0.144 & 4.6 \\

    \bottomrule
    \end{tabular}
\end{center}}\end{minipage}
}

\caption{\textbf{Ablations of Transformer components and the propagation strategy}. Cross-attention contributes most, consistent with the analysis in optical flow task (Table~\ref{tab:transformer}).}
\label{tab:stereo_depth_ablations}

\end{table*}

\begin{table}[t]
    \centering
    \setlength{\tabcolsep}{2pt} %
    \begin{tabular}{llcccccccccccccc}
    \toprule
    
    Training data & Method & EPE & F1-all & $s_{0-10}$ & $s_{10-40}$ & $s_{40+}$ &   \\

    \midrule
    
    \multirow{2}{*}[-2pt]{C + T} & RAFT & 5.32 & 17.46 & 0.67 & 1.58 & 13.68 \\
    & GMFlow & 7.77 & 23.40 & 0.74 & 2.19 & 20.34 \\
    & GMFlow+ & 5.74 & 17.63 & 0.64 & 1.69 & 14.86 \\
    
    \midrule
    
    \multirow{2}{*}[-2pt]{C + T + VK} & RAFT & 2.45 & 7.90 & \textbf{0.43} & 1.18 & 5.70 \\
    & GMFlow & 2.85 & 10.77 & 0.49 & 1.16 & 6.87 \\
    & GMFlow+ & \textbf{2.25} & \textbf{7.20} & 0.48 & \textbf{1.10} & \textbf{5.12} \\

    \bottomrule
    \end{tabular}
    \caption{\textbf{Generalization on KITTI 2015 optical flow dataset} after training on synthetic Chairs (C), Things (T) and Virtual KITTI 2 (VK) datasets. 
    }
    \label{tab:flow_gen_kitti}
    \vspace{-10pt}

\end{table}

\subsubsection{Comparison with RAFT}
\label{sec:compare_raft}

\noindent {\bf Sintel.} Table~\ref{tab:flow_raft_vs_gmflow} shows the results on Things validation set and Sintel (clean and final) training sets after training on Chairs and Things training sets. Without using any refinement, our method achieves better performance on Things and Sintel (clean) than RAFT with 11 refinements. By using an additional task-agnostic hierarchical matching refinement at $1/4$ feature resolution (Sec.~\ref{sec:hierarchical_match}), our method outperforms RAFT with 31 refinements, especially on large motion ($s_{40+}$). Fig.~\ref{fig:flow_iter_vs_epe} visualizes the results. Furthermore, our model enjoys faster inference speed compared to RAFT and also does not require a large number of sequential processing. On the high-end A100 GPU, our model gains more speedup compared to RAFT's sequential architecture ($2.29\times$ \vs. $1.87 \times$, \ie, ours: $151 \to 66$, RAFT: $170 \to 91$), reflecting that our method can benefit more from advanced hardware acceleration and demonstrating its potential for further speed optimization.

\noindent {\bf KITTI.} Table~\ref{tab:flow_gen_kitti} shows the generalization results on KITTI training set after training on Chairs and Things training sets. In this evaluation setting, our method doesn't outperform RAFT, which is mainly caused by the gap between the synthetic training sets and the real-world testing dataset. One key reason behind our inferior performance is that RAFT, relying on fully convolutional neural networks, benefits from the inductive biases in convolution layers, which requires a relatively smaller size training data to generalize to a new dataset in comparison with Transformers~\cite{dosovitskiy2020image, d2021convit,xu2021vitae,zhang2022vitaev2}. To substantiate this claim, we finetune both RAFT and our GMFlow on the additional Virtual KITTI 2~\cite{cabon2020virtual} dataset. The results in Table~\ref{tab:flow_gen_kitti} verify that the performance gap becomes smaller when more data is available. We also train another version GMFlow+ that uses 6 additional local regression refinements (Sec.~\ref{sec:local_reg_refine}), we can observe from Table~\ref{tab:flow_gen_kitti} that GMFlow+ outperforms RAFT on KITTI dataset.

\begin{table}[t]
    \centering
    
    \begin{tabular}{lccccccccccccc}
    \toprule

    \multirow{2}{*}[-2pt]{Attention} & \multicolumn{2}{c}{Things} &
    \multicolumn{2}{c}{KITTI} & \multirow{2}{*}[-2pt]{\begin{tabular}[x]{@{}c@{}}Time\\(ms) \end{tabular}} \\
    \addlinespace[-10pt]  \\
    \cmidrule(lr){2-3} \cmidrule(lr){4-5} 
    \addlinespace[-10pt] \\
    & EPE & D1 & EPE & D1 & \\
    
    \midrule
    
    2D & 1.25 & 3.97 & 1.80 & 13.66 & 61 \\
    1D & \textbf{1.22} & \textbf{3.70} & \textbf{1.61} & \textbf{10.53} & \textbf{50} \\

    \bottomrule
    \end{tabular}
    \caption{\textbf{1D \textit{vs.} 2D cross-attention in Transformer for stereo matching task}. 1D cross-attention is faster and better.}
    \label{tab:stereo_1d_2d_attn}
    \vspace{-10pt}
\end{table}

\subsection{Stereo Matching}

\noindent \textbf{Datasets and evaluation setup.} We first train on the synthetic Scene Flow~\cite{mayer2016large} training set, and then evaluate on the Scene Flow test set and the KITTI 2015~\cite{menze2015object} training set. Unlike previous representative stereo networks~\cite{chang2018pyramid,zhang2019ga,xu2020aanet} that usually rely on a predefined disparity range (typically 192 pixels) to construct the local cost volume, our method is more flexible and can support unconstrained disparity prediction. To avoid extremely large disparity values in the data, we mask the pixels whose disparities exceed 400 pixels during both training and evaluation. Finally, we perform finetuning on KITTI 2015 Stereo, Middlebury Stereo, Argoverse Stereo and ETH3D Stereo datasets and report the performance on the online benchmarks.

\noindent \textbf{Metrics.} We adopt the commonly used metrics end-point-error (EPE) and D1-all, where EPE is the average $\ell_1$ distance between the prediction and ground truth disparity, and D1-all denotes the percentage of outliers.

\noindent \textbf{Training schedule.} For ablations, we train our model on the Scene Flow dataset for 50K iterations, with a batch size of 64 and a learning rate of 1e-3. The finetuning process on each benchmark dataset will be elaborated in Sec.~\ref{sec:stereo_benchmark}.

\subsubsection{Ablations}
\label{sec:stereo_ablation}

\noindent {\bf Stereo cross-attention: 1D \vs \ 2D.} Unlike 2D optical flow, rectified stereo matching is a 1D correspondence task that corresponding pixels lie on the same horizontal scanline. Thus, it's not necessary to perform 2D cross-attention in the Transformer to model cross-view interactions and 1D horizontal cross-attention is sufficient. As shown in Table~\ref{tab:stereo_1d_2d_attn}, using 1D cross-attention is not only more efficient in terms of inference time (measured for KITTI resolution ($384\times 1248$) on a single V100 GPU), but also leads to better performance since unnecessary matching information is avoided. We note that the parameter-free cross-attention operation (2D, 1D or any other forms) doesn't affect the learnable parameters (\ie, the linear projection layers) of the Transformer, and thus the pretrained model for optical flow and stereo matching tasks can still be shared.

\begin{table}[t]
    \centering

    \begin{tabular}{lccccccccccccc}
    \toprule

    \multirow{2}{*}[-2pt]{Model} & \multirow{2}{*}[-2pt]{\#refine} &
    \multirow{2}{*}[-2pt]{EPE} & \multirow{2}{*}[-2pt]{D1} & \multirow{2}{*}[-2pt]{\begin{tabular}[x]{@{}c@{}}Param\\(M) \end{tabular}} & \multirow{2}{*}[-2pt]{\begin{tabular}[x]{@{}c@{}}Time\\(ms) \end{tabular}} \\
    \\
    \midrule

    \multirow{5}{*}[-2pt]{RAFT-Stereo~\cite{lipson2021raft}} & 0 & 3.28 & 13.13 & \multirow{5}{*}[-2pt]{11.1} & 27 \\
    & 3 & 1.20 & 4.50 &  & 36 \\
    & 7 & 0.95 & 3.50 &  & 48 \\
    & 15 & 0.89 & 3.22 & & 73 \\
    & 31 & 0.86 & 3.16 &  & 122 \\
    
    \midrule

    \multirow{3}{*}[-2pt]{\begin{tabular}[x]{@{}l@{}}GMStereo\\(random init) \end{tabular}} & 0 & 1.11 & 3.05 & 4.7 & 23 \\
    & 1 & 0.94 & 2.95 & 4.7 & 58 \\
    & 4 & 0.77 & 2.22 & 7.4 & 86 \\

    \midrule

    \multirow{3}{*}[-2pt]{\begin{tabular}[x]{@{}l@{}}GMStereo\\(flow init) \end{tabular}} & 0 & 1.00 & 2.77 & 4.7 & 23 \\
    & 1 & 0.89 & 2.64 & 4.7 & 58 \\
    & 4 & \textbf{0.72} & \textbf{2.08} & 7.4 & 86 \\

    \bottomrule
    \end{tabular}
    
    \caption{\red{\textbf{Comparison with RAFT-Stereo for stereo matching task}. Our GMStereo trained with random initialization (random init) already significantly outperforms RAFT-Stereo. Leveraging the pretrained GMFlow model as initialization (flow init) makes the performance gap even larger.}}
    \label{tab:compare_raftstereo}
\end{table}

\noindent {\bf Model components.} We ablate different components of our full model in Table~\ref{tab:stereo_ablation}. The results are consistent with those for the optical flow task in Table~\ref{tab:transformer} and Table~\ref{tab:prop}. That is, the cross-attention contributes most, but the other components also contribute to the performance gains.

\subsubsection{\red{Comparison with RAFT-Stereo}}
\label{sec:compare_raftstereo}

\red{
We compare our GMStereo model with RAFT-Stereo~\cite{lipson2021raft} on the Scene Flow test set in Table~\ref{tab:compare_raftstereo}. The prediction error and inference time for different number of refinement steps are reported. We can observe that our GMStereo model trained with random initialization (random init) already significantly outperforms RAFT-Stereo, while having less parameters and running faster. This result is consistent with the comparisons between our GMFlow and RAFT for the optical flow task in Table~\ref{tab:flow_raft_vs_gmflow}. Moreover, our GMStereo model can further benefit from the pretrained flow model thanks to our unified model. As shown in Table~\ref{tab:compare_raftstereo}, our GMStereo model trained with GMFlow model as initialization (flow init) leads to further performance boost, outperforming RAFT-Stereo by even larger margins.
}

\subsection{Depth Prediction}

\noindent \textbf{Datasets and evaluation setup.} For ablations, we train on the ScanNet~\cite{dai2017scannet} dataset, where we follow BA-Net~\cite{Tang2019BANetDB} for the training and testing splits. Finally, we train and evaluate on the SUN3D~\cite{xiao2013sun3d}, RGBD-SLAM~\cite{sturm2012benchmark} and Scenes11~\cite{ummenhofer2017demon} datasets for comparison with previous methods.

\noindent \textbf{Metrics.} Following previous methods~\cite{Tang2019BANetDB, im2019dpsnet}, we use 4 error metrics for evaluation of the depth quality, including Absolute Relative difference (Abs Rel), Squared Relative difference (Sq Rel), Root Mean Squared Error (RMSE) and RMSE in log scale (RMSE log).

\noindent \textbf{Training schedule.} 
For ablation experiments, we train our model on the ScanNet for 50K iterations, with a batch size of 80 and a learning rate of 4e-4. The depth range for training and testing is set to $[0.5, 10]$ meters, and the number of depth candidates in the matching layer (Eq.~\eqref{eq:depth_match}) is set to $64$. The training process on SUN3D~\cite{xiao2013sun3d}, RGBD-SLAM~\cite{sturm2012benchmark} and Scenes11~\cite{ummenhofer2017demon} datasets will be elaborated in Sec.~\ref{sec:depth_benchmark}.

\subsubsection{Ablations}
\label{sec:depth_ablation}

\noindent {\bf Model components.} We ablate different components of our full model in Table~\ref{tab:depth_ablation}. The results are consistent with those for optical flow and stereo matching tasks in Table~\ref{tab:transformer}, Table~\ref{tab:prop} and Table~\ref{tab:stereo_ablation}. That is, the cross-attention contributes most, and other components also contribute to the performance gains.

\begin{table}[t]
    \centering
    \setlength{\tabcolsep}{2.pt} %
    
    \begin{tabular}{lccccccccccccc}
    \toprule

    \multirow{2}{*}[-2pt]{Model} & \multirow{2}{*}[-2pt]{Abs Rel} &
    \multirow{2}{*}[-2pt]{Sq Rel} & \multirow{2}{*}[-2pt]{RMSE} & \multirow{2}{*}[-2pt]{RMSE log} & \multirow{2}{*}[-2pt]{\begin{tabular}[x]{@{}c@{}}Param\\(M) \end{tabular}} & \multirow{2}{*}[-2pt]{\begin{tabular}[x]{@{}c@{}}Time\\(ms) \end{tabular}} \\
    & & & & & & & & & \\

    \midrule

    DeFiNe~\cite{guizilini2022depth} & 0.056 & 0.019 & 0.176 & - & 30.8 & 78 \\
    GMDepth & 0.059 & 0.019 & 0.179 & 0.082 & 7.3 & 40 \\
    
    \bottomrule
    \end{tabular}
    
    \caption{\red{\textbf{Comparison with DeFiNe (Depth Field Network) on ScanNet test set}. DeFiNe relies on a series of 3D geometric augmentations to achieve competitive performance, while our GMDepth can be trained well without any such augmentations. Our method also has $4\times$ less parameters and is $2\times$ faster.}}
    \label{tab:compare_depth_filed_network}
\end{table}

\begin{table}[t]
    \centering
    \setlength{\tabcolsep}{2pt} %
    
    \begin{tabular}{lccccccccccccc}
    \toprule

    \multirow{2}{*}[-2pt]{Model} & \multirow{2}{*}[-2pt]{Abs Rel} &
    \multirow{2}{*}[-2pt]{Sq Rel} & \multirow{2}{*}[-2pt]{RMSE} & \multirow{2}{*}[-2pt]{RMSE log} & \multirow{2}{*}[-2pt]{\begin{tabular}[x]{@{}c@{}}Param\\(M) \end{tabular}} & \multirow{2}{*}[-2pt]{\begin{tabular}[x]{@{}c@{}}Time\\(ms) \end{tabular}} \\
    & & & & & & & & & \\
    
    \midrule

    DepthFormer~\cite{guizilini2022multi} & 0.075 & 0.029 & 0.230 & 0.106 & 5.4 & 29 \\
    GMDepth & \textbf{0.069} & \textbf{0.025} &  \textbf{0.211} &  \textbf{0.097} & \textbf{4.7} & \textbf{17} \\
    
    \bottomrule
    \end{tabular}
    
    \caption{\red{\textbf{Comparison with DepthFormer on ScanNet test set}. Our approach performs better and is more efficient.}}
    \label{tab:compare_depthformer}
\end{table}

\subsubsection{\red{Comparison with Depth Field Network}}

\red{
The Depth Field Network (DeFiNe)~\cite{guizilini2022depth} proposes an implicit way for learning cross-view correspondences, where the geometric priors (\eg, camera information) are encoded as inputs to a Transformer model for depth estimation. Different from DeFiNe, we learn task-agnostic features and obtain the depth prediction with a parameter-free matching layer. In Table~\ref{tab:compare_depth_filed_network}, we show a comparison with DeFiNe on ScanNet test set. Our GMDepth model achieves similar performance but has $4\times $ less parameters and is $2\times$ faster. It is also worth noting that DeFiNe relies on a series of geometric 3D augmentations (\eg, camera transformations) to achieve competitive performance. For example, its `Abs Rel' error increases from 0.093 to 0.117 when such augmentations are removed according to the ablation study in DeFiNe's paper. In contrast, our model can be trained well without any such augmentations. Compared to learning correspondences implicitly like DeFiNe~\cite{guizilini2022depth}, our explicit approach is easier to learn and is more efficient in terms of model parameters and inference speed, since we model the geometric constraints explicitly and the model doesn't need to learn such geometric priors from data.
}

\subsubsection{\red{Comparison with DepthFormer}}

\red{
DepthFormer~\cite{guizilini2022multi} proposes to use a Transformer to improve the quality of cost volume, while we leverage a Transformer to learn strong features for simple parameter-free matching. We compare with DepthFormer in Table~\ref{tab:compare_depthformer} by replacing our Transformer and matching layers with DepthFormer's Transformer-enhanced cost volume and depth decoding layers. We train this model variant within our architecture and keep other components exactly the same. We can observe that our approach performs better. Besides, since DepthFormer's Transformer is applied to the 3D cost volume, which is more computationally expensive than ours that operates on 2D features. Thus our method is also $1.7\times$ faster.
}

\begin{table*}[t]
\centering
\subfloat[
\textbf{Flow to stereo transfer: error curves of disparity prediction error \vs \ numbers of training steps}.
\label{fig:stereo_randinit_flowinit}
]{
\centering
\begin{minipage}{0.42\linewidth}
\includegraphics[width=\linewidth]{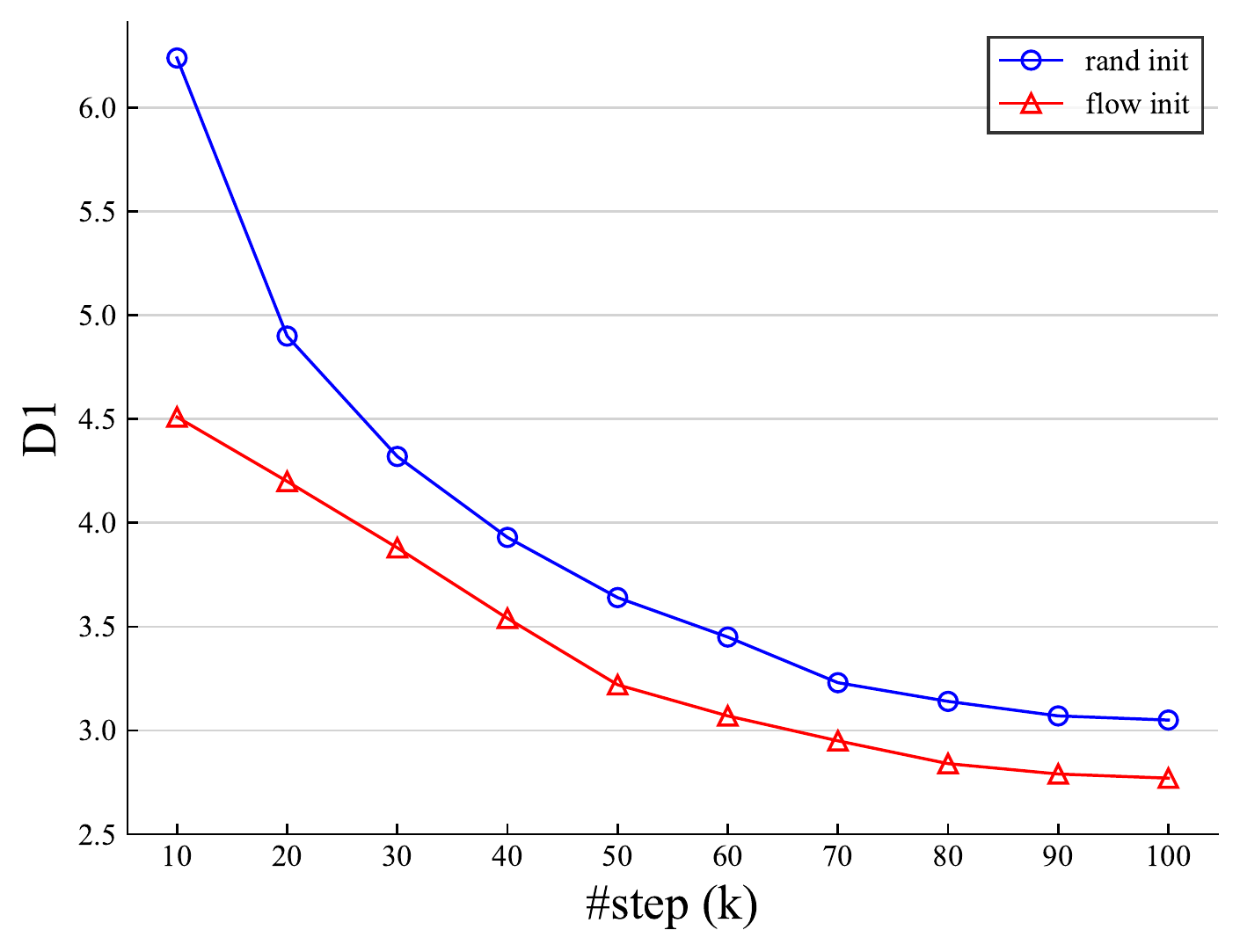}
\end{minipage}
}
\hspace{2em}
\subfloat[
\textbf{Flow to depth transfer: error curves of depth prediction error \vs \ numbers of training steps}.
\label{fig:depth_randinit_flowinit}
]{
\begin{minipage}{0.42\linewidth}
\includegraphics[width=\linewidth]{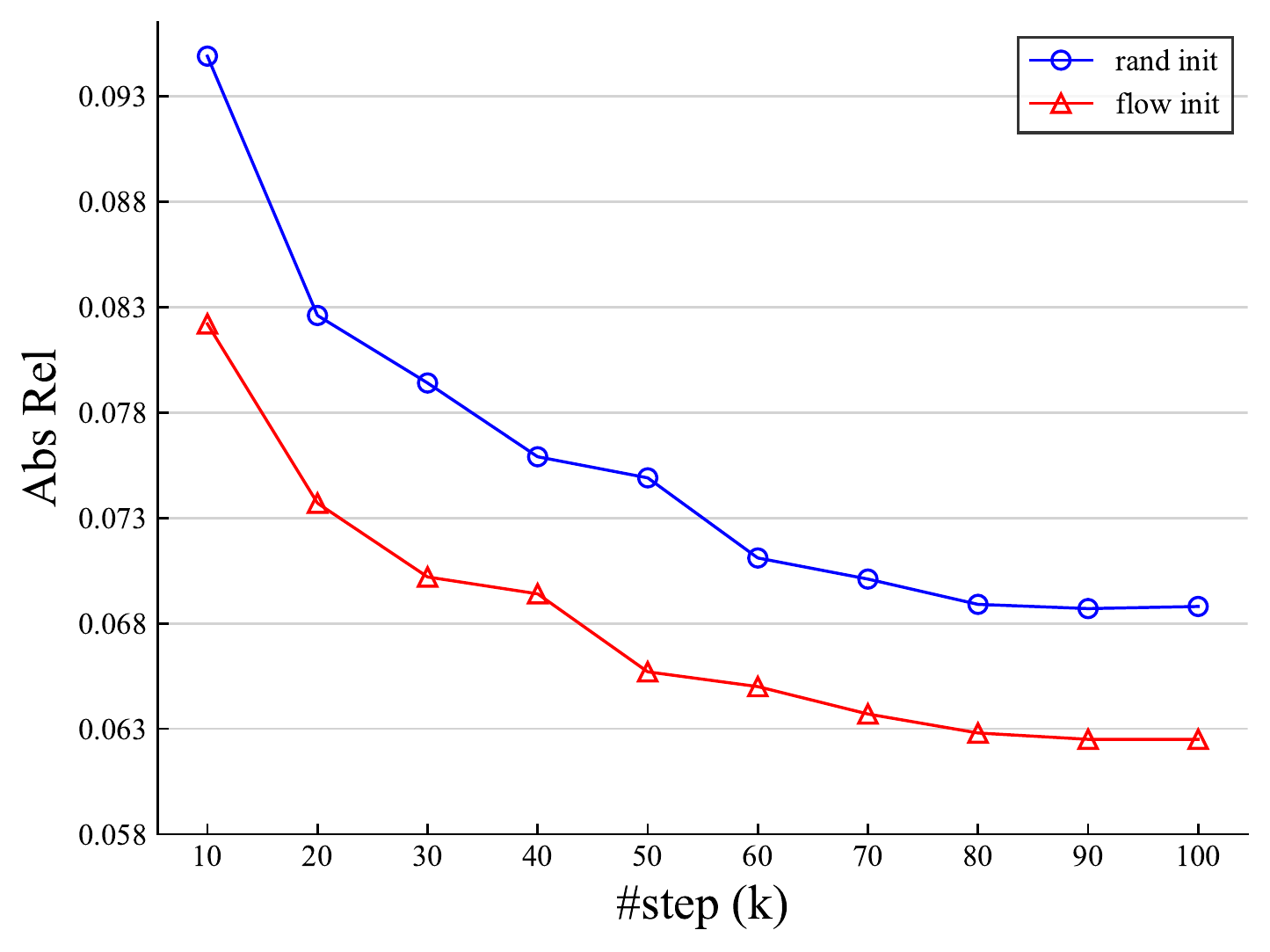}
\end{minipage}
}
\\
\subfloat[
\textbf{Flow to stereo transfer: performance comparison}.
\label{tab:transfer_flow_to_stereo}
]{
\begin{minipage}{0.45\linewidth}{\begin{center}
\begin{tabular}{lccccccccccccc}
\\
    \toprule
    
    \multirow{2}{*}[-2pt]{Model } & \multicolumn{2}{c}{Things} &
    \multicolumn{2}{c}{KITTI} \\
    \addlinespace[-10pt]  \\
    \cmidrule(lr){2-3} \cmidrule(lr){4-5} 
    \addlinespace[-10pt] \\
    & EPE & D1 & EPE & D1 \\

    \midrule
    
    rand init, w/o ft & 75.24 & 96.06 & 103.47 & 96.95 \\
    flow init, w/o ft & 2.58 & 18.19 & 1.98 & 17.60 \\
    
    \midrule
    
    rand init, ft (50K) & 1.22 & 3.70 & 1.61 & 10.53 \\
    flow init, ft (50K) & 1.10 & 3.04 & 1.39 & 7.56 \\
    
    rand init, ft (100K) & 1.11 & 3.05 & 1.58 & 9.93 \\
    flow init, ft (100K) & \textbf{1.00} & \textbf{2.77} & \textbf{1.37} & \textbf{7.38} \\

    \bottomrule
\end{tabular}
\end{center}}\end{minipage}
}
\hspace{2em}
\subfloat[
\textbf{Flow to depth transfer: performance comparison}.
\label{tab:transfer_flow_to_depth}
]{
\begin{minipage}{0.45\linewidth}{\begin{center}
\begin{tabular}{lccccccccccccc}
    \toprule

    Model & Abs Rel & Sq Rel & RMSE & RMSE log \\
    
    \midrule
    
    rand init, w/o ft & 0.536 & 1.309 & 1.300 & 0.584 \\
    flow init, w/o ft & 0.198 & 0.364 & 0.599 & 0.235 \\
    
    \midrule
    
    rand init, ft (50K) & 0.074 & 0.028 & 0.225 & 0.103 \\
    flow init, ft (50K) & 0.066 & 0.023 & 0.203 & 0.092 \\
    
    rand init, ft (100K) & 0.069 & 0.025 & 0.211 & 0.097 \\
    flow init, ft (100K) & \textbf{0.063} & \textbf{0.021} & \textbf{0.193} & \textbf{0.088} \\

    \bottomrule
\end{tabular}
\end{center}}\end{minipage}
}
\\

\caption{\textbf{Cross-task transfer}. We show the comparisons of error curves between random initialization and using a pretrained optical flow model as initialization in Fig.~\ref{fig:stereo_randinit_flowinit} and Fig.~\ref{fig:depth_randinit_flowinit}. The performance comparisons of different models: without any finetuning or finetuned with different initialization (rand init \vs \ flow init) and different numbers of total training steps (50K \vs \ 100K) are shown in Table~\ref{tab:transfer_flow_to_stereo} and Table~\ref{tab:transfer_flow_to_depth}.}
\label{tab:cross_task_transfer}
\end{table*}

\begin{figure}[t]
    \centering
    \includegraphics[width=\linewidth]{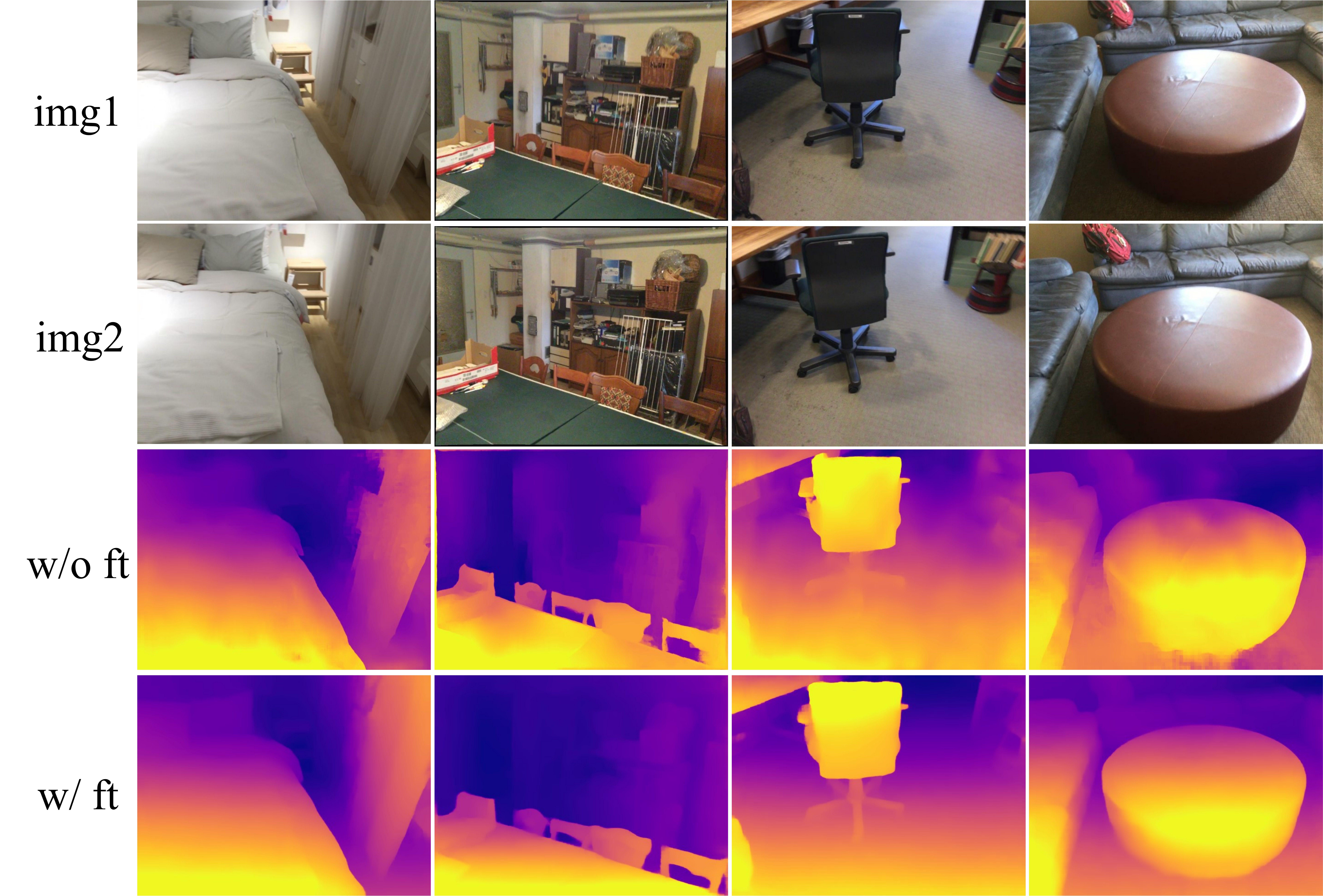}
    \caption{\textbf{Flow to depth transfer}. We use an optical flow model pretrained on Chairs and Things datasets to directly predict depth on the ScanNet dataset, without any finetuning. The performance can be further improved by finetuning on the ScanNet dataset for the depth task.}
    \label{fig:vis_flow2depth}
    \vspace{-10pt}

\end{figure}

\begin{table}[t]
    \centering

    \begin{tabular}{lccccccccccccc}
    \toprule

    \multirow{2}{*}[-2pt]{Model} & \multirow{2}{*}[-2pt]{\#refine} &
    \multirow{2}{*}[-2pt]{EPE} & \multirow{2}{*}[-2pt]{D1} & \multirow{2}{*}[-2pt]{\begin{tabular}[x]{@{}c@{}}Time\\(ms) \end{tabular}} \\
    \\
    \midrule

    \multirow{5}{*}[-2pt]{\begin{tabular}[x]{@{}l@{}}RAFT~\cite{teed2020raft}\\(disparity from $x$-flow) \end{tabular}} & 0 & 8.10 & 34.12 & 15 \\
    & 3 & 2.76 & 8.07 & 22 \\
    & 7 & 2.08 & 6.33 & 31 \\
    & 11 & 1.95 & 6.03 & 41 \\
    & 31 & 1.93 & 5.90 & 95 \\
    
    \midrule

    \begin{tabular}[x]{@{}l@{}}GMFlow\\(1D cross-attention, 1D matching) \end{tabular} & 0 & 2.58 & 18.2 & 23 \\
    \begin{tabular}[x]{@{}l@{}}GMFlow\\(1D cross-attention, 1D matching) \end{tabular} & 1 & 1.38 & 5.27 & 58 \\
    
    \midrule

    \begin{tabular}[x]{@{}l@{}}GMStereo, finetune\\(1D cross-attention, 1D matching) \end{tabular} & 0 & 1.00 & 2.77 & 23 \\
    \begin{tabular}[x]{@{}l@{}}GMStereo, finetune\\(1D cross-attention, 1D matching) \end{tabular} & 1 & \textbf{0.89} & \textbf{2.64} & 58 \\

    \bottomrule
    \end{tabular}
    
    \caption{\red{\textbf{Flow to stereo transfer comparison with RAFT}. The evaluations are conducted on Scene Flow test set for stereo matching task. For RAFT, we obtain the disparity from the $x$ component of its 2D optical flow prediction. Our results are obtained by modifying the cross-attention function and the matching layer.}}
    \label{tab:flow2stereo_compare_raft}
    \vspace{-10pt}
\end{table}

\begin{table}[t]
    \centering
    \setlength{\tabcolsep}{2.pt} %

    \begin{tabular}{lccccccccccccc}
    \toprule

    Model & \#refine & Abs Rel & Sq Rel & RMSE & RMSE log \\
    \\
    \midrule

    \multirow{5}{*}[-2pt]{\begin{tabular}[x]{@{}l@{}}RAFT~\cite{teed2020raft}\\(depth from triangulation) \end{tabular}} & 0 & 0.271 & 0.359 & 0.781 & 1.062 \\
    & 3 & 0.146 & 0.144 & 0.464 & 0.428 \\
    & 7 & 0.123 & 0.110 & 0.401 & 0.302 \\
    & 11 & 0.118 & 0.103 & 0.388 & 0.283 \\
    & 31 & 0.117 & 0.102 & 0.385 & 0.271 \\
    
    \midrule

    \begin{tabular}[x]{@{}l@{}}GMFlow\\(depth matching) \end{tabular} & 0 & 0.198 & 0.364 & 0.599 & 0.235 \\

    \midrule

    \begin{tabular}[x]{@{}l@{}}GMDepth, finetune\\(depth matching) \end{tabular} & 0 & \textbf{0.063} & \textbf{0.021} & \textbf{0.193} & \textbf{0.088} \\

    \bottomrule
    \end{tabular}
    
    \caption{\red{\textbf{Flow to depth transfer comparison with RAFT}. The evaluations are conducted on ScanNet test set. For RAFT, we compute the optical flow first and then obtain the depth prediction with triangulation. Our results are obtained by modifying the matching layer.}}
    \label{tab:flow2depth_compare_raft}
\end{table}

\subsection{Cross-Task Transfer}

One unique benefit of our unified model is that it naturally enables cross-task transfer since all the learnable parameters are exactly the same. More specifically, we can directly use a pretrained optical flow model and apply it to both rectified stereo matching and unrectified stereo depth estimation tasks. As shown in Table~\ref{tab:transfer_flow_to_stereo} and Table~\ref{tab:transfer_flow_to_depth}, our pretrained optical flow model performs significantly better than a random initialized model. The visual results are shown in Fig.~\ref{fig:vis_flow2depth}, where  our model achieves promising results. The pretrained flow model can be further finetuned for stereo and depth tasks, which not only leads to faster training speed, but also achieves better performance than random initialization (Table~\ref{tab:cross_task_transfer}).

We also experiment with transferring the pretrained models from the stereo and depth tasks to the optical flow task, but no obvious performance gain is observed. This is understandable since stereo and depth are both 1D correspondence problems, and their pretrained models might be specialized to the 1D correspondence matching task and thus are not able to bring clear benefits to the more general 2D correspondence task (\ie, optical flow).

\subsubsection{\red{Flow to Stereo Transfer Comparison with RAFT}}

\red{
We compare with RAFT in terms of flow to stereo transfer in Table~\ref{tab:flow2stereo_compare_raft}. More specifically, we use RAFT to extract optical flow from a stereo pair and obtain the disparity from the horizontal component of the 2D optical flow. For our method, we are able to obtain the disparity from our flow model GMFlow by modifying the parameter-free cross-attention function and the matching layer. We can observe from Table~\ref{tab:flow2stereo_compare_raft} that our GMFlow with only 1 refinement already outperforms RAFT with 31 refinements, without any finetuning for the stereo task. Our unified model is able to benefit from additional finetuning and achieves further performance improvement.
}

\begin{table}[t]
    \centering
    \setlength{\tabcolsep}{2pt} %
    \begin{tabular}{lccccccccccccccr}
    \toprule
    
    \multirow{2}{*}[-2pt]{Method} & \multirow{2}{*}[-2pt]{\#refine.} & \multicolumn{4}{c}{Things (val, clean)} & \multicolumn{1}{c}{Sintel (clean)}  \\
    \addlinespace[-10pt] \\
    \cmidrule(lr){3-6} \cmidrule(lr){7-7} 
    \addlinespace[-10pt] \\
    & & EPE & $s_{0-10}$ & $s_{10-40}$ & $s_{40+}$ & EPE & \\
    
    \midrule
    
    GMFlow & 1 & 2.80 & 0.53 & 1.01 & 7.31 & 1.08 \\
    
    \midrule 
    
    \multirow{3}{*}[-2pt]{{GMFlow+}} & 2 & 2.52 & 0.46 & 0.88 & 6.63 & 1.03 \\
    & 4 & 2.29 & 0.34 & 0.75 & 6.16 & 0.94 \\
    & 7 & \textbf{2.20} & \textbf{0.30} & \textbf{0.69} & \textbf{5.97} & \textbf{0.91} \\

    \bottomrule
    \end{tabular}
    \caption{\textbf{Additional local regression refinements for optical flow task}. 
    }
    \label{tab:flow_more_refine}
\end{table}

\begin{table}[t]
\footnotesize
    \centering
    \setlength{\tabcolsep}{2pt} %
    \begin{tabular}{lcccccc}
    \toprule
    \multirow{2}{*}[-2pt]{Method } & \multicolumn{3}{c}{Sintel (clean)} &
    \multicolumn{3}{c}{Sintel (final)} \\
    \addlinespace[-10pt]  \\
    \cmidrule(lr){2-4} \cmidrule(lr){5-7} 
    \addlinespace[-10pt] \\
    & all & matched & unmatched & all & matched & unmatched \\
    \midrule
    
    FlowNet2~\cite{ilg2017flownet} & 4.16 & 1.56 & 25.40 & 5.74 & 2.75 & 30.11 \\
    PWC-Net+~\cite{sun2019models} & 3.45 & 1.41 & 20.12 & 4.60 & 2.25 & 23.70  \\
    HD$^3$~\cite{yin2019hierarchical} & 4.79 & 1.62 & 30.63 & 4.67 & 2.17 & 24.99 \\
    VCN~\cite{yang2019volumetric} & 2.81 & 1.11 & 16.68 & 4.40 & 2.22 & 22.24 \\
    DICL~\cite{wang2020displacement} & 2.63 & 0.97 & 16.24 & 3.60 & 1.66 & 19.44 \\
    
    RAFT~\cite{teed2020raft} & 1.94 & - & - & 3.18 & - & - \\
    GMFlow~\cite{xu2022gmflow} & 1.74 & 0.65 & 10.56 & 2.90 & 1.32 & 15.80 \\

    RAFT$^\dagger$~\cite{teed2020raft} & 1.61 & 0.62 & 9.65 & 2.86 & 1.41 & 14.68 \\
    GMA$^\dagger$~\cite{Jiang_2021_ICCV} & 1.39 & 0.58 & 7.96 & 2.47 & 1.24 & 12.50 \\
    GMFlowNet~\cite{zhao2022global} & 1.39 & 0.52 & 8.49 & 2.65 & 1.27 & 13.88 \\
    DIP$^\dagger$~\cite{zheng2022dip} & 1.44 & 0.52 & 8.92 & 2.83 & 1.28 & 15.49 \\ 
    AGFlow$^\dagger$~\cite{luo2022learning} & 1.43 & 0.56 & 8.54 & 2.47 & 1.22 & 12.64 \\
    CRAFT$^\dagger$~\cite{sui2022craft} & 1.44 & 0.61 & 8.20 & 2.42 & 1.16 & 12.64 \\
    FlowFormer~\cite{huang2022flowformer} & 1.20 & 0.41 & 7.63 & \textbf{2.12} & \textbf{0.99} & \textbf{11.37} \\

    GMFlow+ & \textbf{1.03} & \textbf{0.34} & \textbf{6.68} & 2.37 & 1.10 & 12.74 \\

    \bottomrule
    \end{tabular}
    \caption{\textbf{Comparisons on Sintel test test for optical flow.} $^\dagger$ represents the method uses last frame's flow prediction as initialization for subsequent refinement, while other methods all use two frames only.}
    \label{tab:flow_sintel_test}
\end{table}

\begin{figure}[t]
    \centering
    \includegraphics[width=\linewidth]{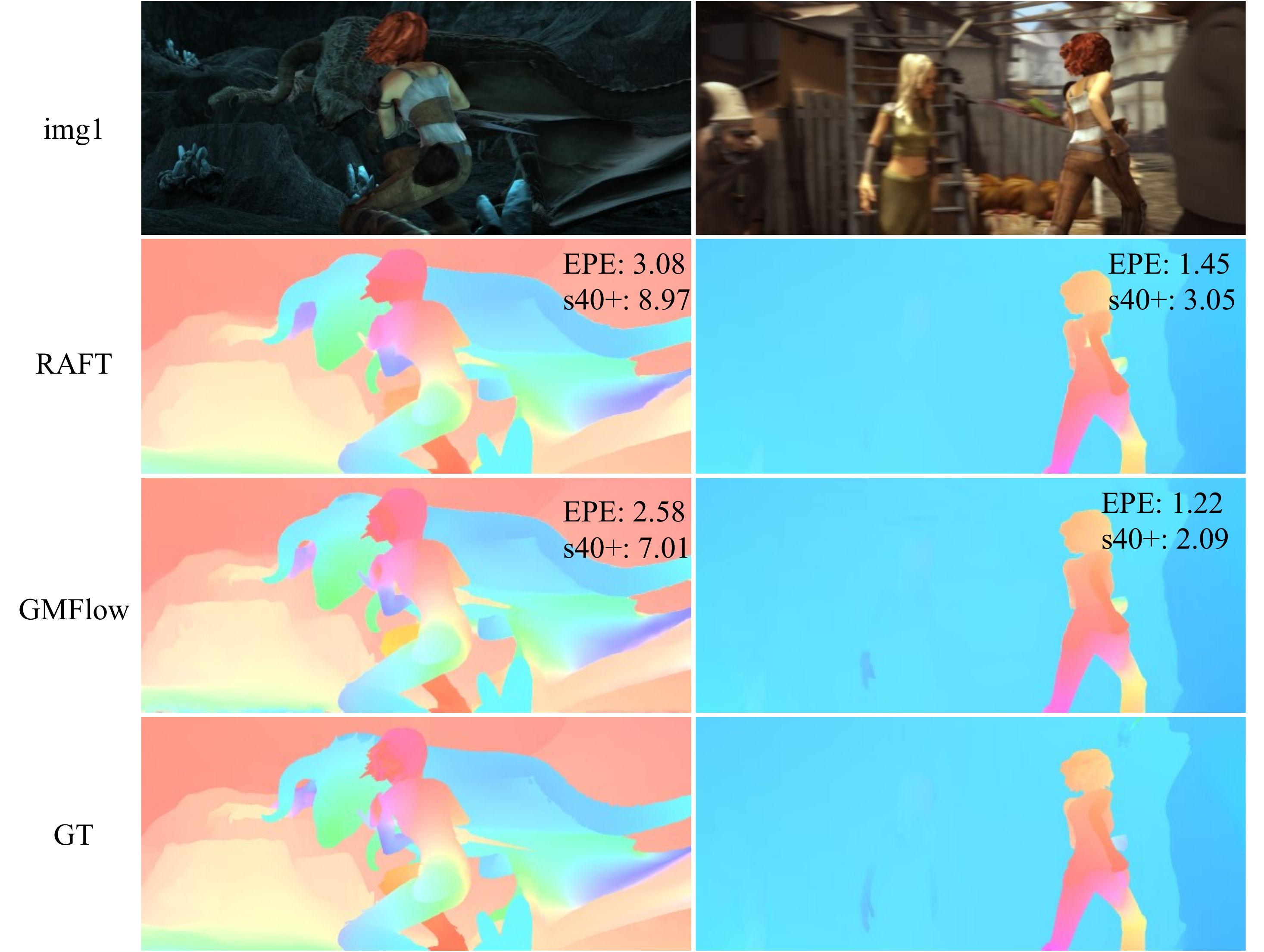}
    \caption{\textbf{Visual comparisons on Sintel test set}.}
    \label{fig:vis_sintel_compare}

\end{figure}

\subsubsection{\red{Flow to Depth Transfer Comparison with RAFT}}

\red{
We compare with RAFT in terms of flow to depth transfer in Table~\ref{tab:flow2depth_compare_raft}. More specifically, we use RAFT to extract optical flow from two posed images and obtain the depth prediction with triangulation~\cite{ke2021deep}. For our method, we obtain the depth prediction by modifying the matching layer to the depth task. We can observe from Table~\ref{tab:flow2depth_compare_raft} that our method performs better than RAFT in terms of the `RMSE log' metric, but inferior for other metrics. This indicates that our results might have large outliers that dominate the averaged metrics of `Abs Rel', `Sq Rel' and `RMSE', but their influence becomes weaker when evaluated in the log scale. The possible reason for this phenomenon is that unlike stereo disparity that is a special case of optical flow, the depth matching layer is slightly different from the flow one, which might make it challenging to do direct cross-task transfer and some outliers might exist. In contrast, the triangulation process involves solving a least-square problem, which is more complex than our simple argmax operation. However, one unique strength of our unified model is that the pretrained flow model can be finetuned for the depth task, and it can quickly adapt to the depth task and finally outperforms the triangulation-based approach by a large margin.
}

\begin{table}[t]
\footnotesize
    \centering
    
    \begin{tabular}{lccccccccccccc}
    \toprule

    Method & Non-occluded pixels & All pixels \\

    \midrule 
    
    FlowNet2~\cite{ilg2017flownet} & 6.94 & 10.41 \\
    PWC-Net+~\cite{sun2019models} & 4.91 & 7.72 \\
    HD$^3$~\cite{yin2019hierarchical} & 3.93 & 6.55 \\
    VCN~\cite{yang2019volumetric} & 3.89 & 6.30 \\
    RAFT~\cite{teed2020raft} & 3.07 & 5.10 \\
    CRAFT~\cite{sui2022craft} & 3.02 & 4.79 \\
    SeparableFlow~\cite{zhang2021separable} & 2.78 & 4.53 \\
    GMFlowNet~\cite{zhao2022global} & 2.75 & 4.79 \\
    DEQ-Flow~\cite{bai2022deep} & 2.96 & 4.91 \\
    AGFlow~\cite{luo2022learning} & 2.97 & 4.89 \\
    KPA-Flow~\cite{luo2022blearning} & 2.82 & 4.60 \\
    FlowFormer~\cite{huang2022flowformer} & 2.69 & 4.68 \\
    GMFlow+ & \textbf{2.40} & \textbf{4.49} \\
    
    \bottomrule
    \end{tabular}
    \caption{\textbf{\red{Comparisons on KITTI test set for optical flow}}.}
    \label{tab:flow_kitti_test}
\end{table}

\begin{table}[t]
    \centering
    
    \begin{tabular}{lccccccccccccc}
    \toprule

    \multirow{2}{*}[-2pt]{setup} & \multirow{2}{*}[-2pt]{\#refine.} &  \multicolumn{2}{c}{Things} &
    \multicolumn{2}{c}{KITTI} \\ %
    \addlinespace[-10pt]  \\
    \cmidrule(lr){3-4} \cmidrule(lr){5-6} 
    \addlinespace[-10pt] \\
    & & EPE & D1 & EPE & D1 \\
    
    \midrule
    
    baseline & 1 & 0.94 & 2.95 & 1.31 & 6.79 \\ %
    
    \midrule
    
    1D correlation & \multirow{2}{*}[-2pt]{2} & 0.84 & 2.46 & 1.27 & 6.22 \\ %
    2D correlation & & \textbf{0.83} & \textbf{2.42} & \textbf{1.25} & \textbf{5.96} \\ %
    
    \midrule
    1D correlation & \multirow{2}{*}[-2pt]{4} & 0.82 & 2.32 & 1.32 & 6.50 \\ %
    2D correlation & & \textbf{0.77} & \textbf{2.22} & \textbf{1.24} & \textbf{6.00} \\ %

    \bottomrule
    \end{tabular}
    \caption{\textbf{Additional local regression refinement for stereo matching task}. We observe that 2D correlation is better than 1D correlation in the local correlation and convolution-based regression method.}
    \label{tab:stereo_refine_1d_2d_corr}
\end{table}

\subsection{Benchmark Results}

In this section, we perform system-level comparisons with previous methods on standard optical flow, stereo matching and depth estimation benchmarks.

\subsubsection{Optical Flow}
\label{sec:flow_benchmark}

In Sec.~\ref{sec:compare_raft}, we have demonstrated that our unified model with 1 additional task-agnostic hierarchical matching refinement at $1/4$ feature resolution can already outperform 31-refinement RAFT. To fully unleash the potential of our method, we use additional task-specific post-processing steps for further improvement. More specifically, we use 6 additional RAFT's iterative local regression refinements at $1/4$ feature resolution, which can further improve our performance on unmatched regions and fine-grained details, as shown in Table~\ref{tab:flow_more_refine}. We note that other post-processing strategies might also be applicable to our method, in this paper we adopt RAFT's approach for convenience.

\begin{table}[!t]
    \centering
    
    \begin{tabular}{lccccccccccccc}
    \toprule

    Model & D1-all (All) & D1-all (Noc) & Time (s) \\
    
    \midrule
    
    LEAStereo~\cite{cheng2020hierarchical} & \textbf{1.65} & \textbf{1.51} & 0.30 \\
    CREStereo~\cite{li2022practical} & 1.69 & 1.54 & 0.41 \\
    GANet-deep~\cite{zhang2019ga} & 1.81 & 1.63 & 1.80 \\
    CFNet~\cite{shen2021cfnet} & 1.88 & 1.73 & 0.18 \\
    AANet+~\cite{xu2020aanet} & 2.03 & 1.85 & 0.06 \\
    PSMNet~\cite{chang2018pyramid} & 2.32 & 2.14 & 0.41 \\
    GMStereo & 1.77 & 1.61 & 0.17 \\

    \bottomrule
    \end{tabular}
    \caption{\textbf{Stereo performance on KITTI 2015 test set}. }
    \label{tab:stereo_kitti15_test}
    \vspace{-10pt}
    
\end{table}

\begin{figure}[t]
    \centering
    \includegraphics[width=\linewidth]{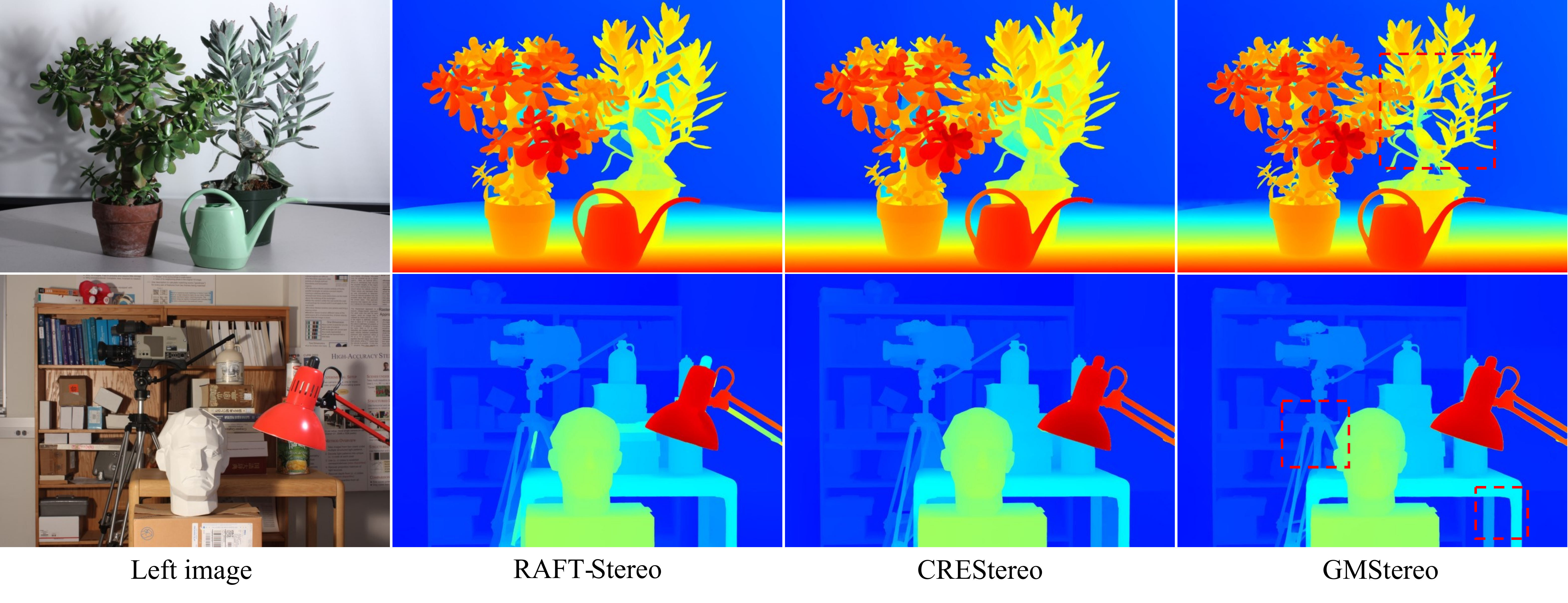}
    \caption{\textbf{Visual comparisons on Middlebury test set}.}
    \label{fig:vis_middlebury_compare}
    \vspace{-10pt}

\end{figure}

\begin{table}[t]
    \centering
    \setlength{\tabcolsep}{4.pt} %
    
    \begin{tabular}{lccccl}
    \toprule
    
    Model & bad 2.0 & bad 4.0 & AvgErr & RMS & Time (s) \\
    
    \midrule
    
    CREStereo~\cite{li2022practical} & \textbf{3.71} & \textbf{2.04} & \textbf{1.15} & 7.70 & 3.55 (F) \\
    RAFT-Stereo~\cite{lipson2021raft} & 4.74 & 2.75 & 1.27 & 8.41 & 11.6 (F) \\
    LEAStereo~\cite{cheng2020hierarchical} & 7.15 & 2.75 & 1.43 & 8.11 & 2.90 (H) \\
    HSMNet~\cite{yang2019hierarchical} &  10.2 & 4.83 & 2.07 & 10.3 & 0.51 (F) \\
    CFNet~\cite{shen2021cfnet} & 10.1 & 6.49 & 3.49 & 15.4 & 0.69 (H) \\
    GMStereo & 7.14 & 2.96 & 1.31 & \textbf{6.45} & 0.73 (F) \\

    \bottomrule
    \end{tabular}
    \caption{\textbf{Stereo performance on Middlebury test set}. ``F'' and ``H'' denote the results are generated using the full and half resolution images, respectively.}
    \label{tab:stereo_middlebury_test}
\end{table}

\noindent \textbf{Sintel.} 
Following RAFT~\cite{teed2020raft}, we further finetune our Things trained model on several mixed datasets that consist of KITTI 2015~\cite{menze2015object}, HD1K~\cite{kondermann2016hci}, FlyingThings3D~\cite{mayer2016large} and Sintel~\cite{butler2012naturalistic} training sets. We first finetune on the mixed dataset for 200K iterations with a batch size of 8 and a learning rate of 2e-4. Then we finetune on the Sintel training sets only with a larger crop size $416 \times 1024$ for 5K iterations. The batch size is 8 and the learning rate is 1e-4. To generate the flow prediction results on the Sintel test sets, we first resize the original images to $416 \times 1024$ and then resize the prediction back to the original image resolution for submission. 
The results on Sintel test set are shown in Table~\ref{tab:flow_sintel_test}. We achieve state-of-the-art results on the highly competitive Sintel (clean) dataset. On Sintel (final) dataset, our performance is only second to the recent FlowFormer~\cite{huang2022flowformer} model, which uses a Transformer model that is pretrained on the large scale ImageNet dataset and is more computationally expensive due to the large number of sequential refinements like RAFT. The visual comparisons with RAFT are shown in Fig.~\ref{fig:vis_sintel_compare}, our method can better capture the motion of fast-moving objects like the moving hand.

\noindent \textbf{KITTI.} 
We further finetune our VKITTI2 trained GMFlow+ model (Table~\ref{tab:flow_gen_kitti}) on the mixed KITTI 2012 and KITTI 2015 training sets for 30K iterations. The batch size is 8 and the learning rate is 2e-4. 
The comparison results with previous methods are shown in Table~\ref{tab:flow_kitti_test}. We outperform all previous methods.

\subsubsection{Stereo Matching}
\label{sec:stereo_benchmark}

Similar to optical flow, we use additional task-specific local regression refinements to further improve our performance. Although the rectified stereo matching is a 1D correspondence task, we found that 2D correlation in the cost volume and convolution-based regression method performs better than 1D correlation (Table~\ref{tab:stereo_refine_1d_2d_corr}), and thus 2D correlation is used in our final model. From Table~\ref{tab:stereo_refine_1d_2d_corr}, we can also observe the performance gets saturated with 3 more local regression refinements, and thus our final model uses 3 additional refinements besides 1 hierarchical matching refinement at $1/4$ feature resolution. {We note again that the number of refinements required by our final model is much smaller than compared to pure iterative architectures like RAFT-Stereo~\cite{lipson2021raft} and CREStereo~\cite{li2022practical} thanks to our strong discriminative feature representations.}

\noindent \textbf{KITTI.} 
We first finetune our Scene Flow pretrained model on the Virtual KITTI 2~\cite{cabon2020virtual} dataset for 30K iterations, with a batch size of 16 and a learning rate of 4e-4. We then further finetune on the mixed KITTI 2012 and KITTI 2015 training sets for 10K iterations, with a batch size of 16 and a learning rate of 4e-4. The final model is used to generate the disparity prediction results on KITTI 2015 test set for submission to the online benchmark. 
The results are shown in Table~\ref{tab:stereo_kitti15_test}. We achieve competitive performance compared with the state-of-the-art methods LEAStereo~\cite{cheng2020hierarchical} and CREStereo~\cite{li2022practical}. Besides, our model runs about $2\times$ faster since we don't rely on any 3D convolutions (unlike LEAStereo) or a large number ($20+$) of sequential refinements (unlike CREStereo). Compared with previous lightweight stereo model AANet~\cite{xu2020aanet}, our method performs much better. Besides, our model can be implemented with pure PyTorch, without requiring to build additional CUDA ops like AANet, which demonstrates that our method achieves a better speed-accuracy trade-off and has more practical advantages.

\begin{table}[!t]
    \centering
    
    \begin{tabular}{lccccccccccccc}
    \toprule
    
    Model & bad 1.0 & bad 2.0 & bad 4.0 \\
    
    \midrule
    GANet~\cite{zhang2019ga} & 6.56 & 1.10 & 0.54 \\
    AANet~\cite{xu2020aanet} & 5.01 & 1.66 & 0.75 \\
    CFNet~\cite{shen2021cfnet} & 3.31 & 0.77 & 0.31 \\
    RAFT-Stereo~\cite{lipson2021raft} & 2.44 & 0.44 & 0.15 \\
    CREStereo~\cite{li2022practical} & \textbf{0.98} & \textbf{0.22} & 0.10 \\
    GMStereo & 1.83 & 0.25 & \textbf{0.08} \\

    \bottomrule
    \end{tabular}
    \caption{\textbf{Stereo performance on ETH3D stereo test set}. }
    \label{tab:stereo_eth3d_test}
\end{table}

\begin{table}[!t]
    \centering
    
    \begin{tabular}{lccccccccccccc}
    \toprule
    
    Model & all:10 & all:5 & all:3 & Time (ms) \\
    
    \midrule
    
    ACVNet~\cite{xu2022attention} & 4.06 & 6.46 & 10.10 & 236 \\
    Odepth$^\dagger$ & 3.78 & 7.55 & 12.33 & 199 \\
    LRM$^\dagger$ & 2.47 & 4.71 & 8.44 & 191 \\
    MSCLab$^\dagger$ & 2.39 & 6.29 & 11.65 & 150 \\
    GMStereo & \textbf{1.61} & \textbf{3.19} & \textbf{6.86} & 190 \\

    \bottomrule
    \end{tabular}
    \caption{\textbf{Argoverse Stereo Challenge on Autonomous Driving Workshop}. $^\dagger$ denotes anonymous submission.}
    \label{tab:stereo_argo_test}
    \vspace{-10pt}
\end{table}

\noindent \textbf{Middlebury.} 
Following CREStereo~\cite{li2022practical}, we collect several public stereo datasets for training. More specifically, we first finetune the Scene Flow pretrained model on the mixed Scene Flow~\cite{mayer2016large}, Tartan Air~\cite{tartanair2020iros}, Falling Things~\cite{tremblay2018falling}, CARLA HR-VS~\cite{yang2019hierarchical}, CREStereo Dataset~\cite{li2022practical}, InStereo2K~\cite{bao2020instereo2k} and Middlebury~\cite{hirschmuller2007evaluation,scharstein2014high} datasets for 100K iterations. The batch size is 16 and the learning rate is 4e-4. We use a random crop size of $480\times 640$ for training at this stage. To adapt the model to high resolution (\eg, $1536\times 2048$ for Middlebury), we perform another round of finetuning with a larger random crop size of $768\times 1024$. The training datasets include CARLA HR-VS~\cite{yang2019hierarchical}, CREStereo Dataset~\cite{li2022practical}, InStereo2K~\cite{bao2020instereo2k}, Falling Things~\cite{tremblay2018falling} and Middlebury~\cite{hirschmuller2007evaluation, scharstein2014high}. At inference, we resize all the Middlebury full resolution test images to $1536 \times 2048$ for prediction and finally resize the predicted disparities back to the original image resolution for evaluation. 
The results are shown in Table~\ref{tab:stereo_middlebury_test}. Our GMStereo achieves the first place in terms of the RMS (Root Mean Square) disparity error metric. Besides, our method shows much higher efficiency than CREStereo~\cite{li2022practical} ($5\times$ faster) and RAFT-Stereo~\cite{lipson2021raft} ($15 \times$ faster) on such a high-resolution dataset. We also show some visual comparisons in Fig.~\ref{fig:vis_middlebury_compare}, our method produces sharper object structures than CREStereo~\cite{li2022practical} and RAFT-Stereo~\cite{lipson2021raft}.

\noindent \textbf{ETH3D.} 
Similar to the training process on the Middlebury dataset, we use several public stereo datasets for training. More specifically, we first finetune the Scene Flow pretrained model on the mixed Scene Flow~\cite{mayer2016large}, Tartan Air~\cite{tartanair2020iros}, Sintel Stereo~\cite{butler2012naturalistic}, CREStereo Dataset~\cite{li2022practical}, InStereo2K~\cite{bao2020instereo2k} and ETH3D~\cite{schops2017multi} datasets for 100K iterations. The batch size is 24 and the learning rate is 4e-4. Then we perform another round of finetuning on the mixed CREStereo Dataset~\cite{li2022practical}, InStereo2K~\cite{bao2020instereo2k} and ETH3D~\cite{schops2017multi} datasets for 30K iterations. Again, the batch size is 24 and the learning rate is 4e-4. 
The results are shown in Table~\ref{tab:stereo_eth3d_test}. We achieve the second place in terms of the `bad 1.0' and `bad 2.0' metrics and the first place in terms of the `bad 4.0' metric.

\noindent \textbf{Argoverse.} We also participate in the Argoverse Stereo Challenge held in the context of the CVPR 2022 Autonomous Driving Workshop\footnote{\url{http://cvpr2022.wad.vision/}} to further demonstrate the potential of our method. Since this competition requires algorithms to produce disparity predictions in 200ms or less, we use global matching at $1/8$ feature resolution and two additional local regression refinements also at $1/8$ resolution. The inference time is about 190ms for a stereo pair of $1024 \times 1232$ resolution (resized from the full resolution for prediction). 
We first train our model on the Scene Flow dataset with GMFlow’s FlyingChairs and FlyingThings3D pretrained model as initialization. Then, we finetune on the mixed Virtual KITTI 2~\cite{cabon2020virtual} and DrivingStereo~\cite{yang2019drivingstereo} datasets. Finally, we finetune on the mixed Argoverse training and validation splits and the final disparity prediction results are submitted to the online test server for evaluation. 
The results are shown in Table~\ref{tab:stereo_argo_test}, where our GMStereo achieves the first place and clearly outperforms all other submissions.

\begin{table}[t]
    \centering
    \setlength{\tabcolsep}{3.pt} %
    
    \begin{tabular}{lccccccccccccc}
    \toprule

    Model & Abs Rel & Sq Rel & RMSE & RMSE log & Time (s) \\
    
    \midrule
    
    DeMoN~\cite{ummenhofer2017demon} & 0.231 & 0.520 & 0.761 & 0.289 & 0.69 \\
    BA-Net~\cite{Tang2019BANetDB} & 0.161 & 0.092 & 0.346 & 0.214 & 0.38 \\
    DeepV2D~\cite{Teed2020DeepV2D} & \textbf{0.057} & \textbf{0.010} & \textbf{0.168} & \textbf{0.080} & 0.69 \\
    GMDepth & 0.059 & 0.019 & 0.179 & 0.082 & 0.04 \\

    \bottomrule
    \end{tabular}
    \caption{\textbf{Depth performance on ScanNet test set}. }
    \label{tab:depth_scannet_test}
\end{table}

\begin{table}[t]
    \centering
    \setlength{\tabcolsep}{3.pt} %
    
    \begin{tabular}{llcccccccccccc}
    \toprule
    
    Dataset & Model & Abs Rel & Sq Rel & RMSE & RMSE log \\
    
    \midrule
    
    \multirow{5}{*}[-2pt]{RGBD-SLAM} & DeMoN~\cite{ummenhofer2017demon} & 0.157 & 0.524 & 1.780 & 0.202  \\
    & DeepMVS~\cite{huang2018deepmvs} & 0.294 & 0.430 & 0.868 & 0.351 \\
    & DPSNet~\cite{im2019dpsnet} & 0.154 & 0.215 & 0.723 & 0.226 &   \\
    & IIB~\cite{yifan2022input} & \textbf{0.095} & - & \textbf{0.550} & - &  \\
    & GMDepth & 0.101 & \textbf{0.177} & 0.556 & \textbf{0.167} \\
    
    \midrule
    
    \multirow{5}{*}[-2pt]{SUN3D} & DeMoN~\cite{ummenhofer2017demon} & 0.214 & 1.120 & 2.421 & 0.206  \\
    & DeepMVS~\cite{huang2018deepmvs} & 0.282 & 0.435 & 0.944 & 0.363 \\
    & DPSNet~\cite{im2019dpsnet} & 0.147 & 0.107 & 0.427 & 0.191 \\
    & IIB~\cite{yifan2022input} & \textbf{0.099} & - & \textbf{0.29}3 & - &  \\
    & GMDepth & 0.112 & \textbf{0.068} & 0.336 & \textbf{0.146} \\
    
    \midrule
    
    \multirow{5}{*}[-2pt]{Scenes11} & DeMoN~\cite{ummenhofer2017demon} & 0.556 & 3.402 & 2.603 & 0.391 \\
    & DeepMVS~\cite{huang2018deepmvs} & 0.210 & 0.373 & 0.891 & 0.270 \\
    & DPSNet~\cite{im2019dpsnet} & 0.056 & 0.144 & 0.714 & 0.140 \\
    & IIB~\cite{yifan2022input} & 0.056 & - & 0.523 & - \\
    & GMDepth & \textbf{0.050} & \textbf{0.069} & \textbf{0.491} & \textbf{0.106} \\

    \bottomrule
    \end{tabular}
    \caption{\textbf{Depth performance on RGBD-SLAM, SUN3D and Scenes11 test datasets}. }
    \vspace{-10pt}
    \label{tab:depth_demon_test}
\end{table}

\subsubsection{Depth Estimation}
\label{sec:depth_benchmark}

For unrectified two-view depth estimation, our final model doesn't use the hierarchical matching refinement at $1/4$ feature resolution since we don't observe very large performance gains. Instead, we use an additional task-specific local regression refinement at $1/8$ feature resolution, which further improves the performance while maintaining fast inference speed.

\noindent \textbf{ScanNet.} 
Following BA-Net~\cite{Tang2019BANetDB}'s training and testing splits on ScanNet~\cite{dai2017scannet}, we train our GMDepth model on the training split for 100K iterations. The batch size is 80 and the learning rate is 4e-4. 
The results are shown in Table~\ref{tab:depth_scannet_test}. We achieve performance comparable to the representative method DeepV2D~\cite{Teed2020DeepV2D} and outperform DeMoN~\cite{ummenhofer2017demon} and BA-Net~\cite{Tang2019BANetDB} by a large margin. Notably, our model runs $~10 \times$ faster than BA-Net and $~15 \times$ faster than DeepV2D, which both heavily rely on computationally expensive 3D convolutions. This demonstrates that our model has strong potential for real-world use cases.

\noindent \textbf{RGBD-SLAM, SUN3D, and Scenes11.} 
Following the setting of DPSNet~\cite{im2019dpsnet}, we train our model on the mixed RGBD-SLAM~\cite{sturm2012benchmark}, SUN3D~\cite{xiao2013sun3d} and Scenes11~\cite{ummenhofer2017demon} training sets for 100K iterations, with a batch size of 64 and a learning rate of 4e-4.
The evaluation results on respective RGBD-SLAM, SUN3D and Scenes11 test sets are shown in Table~\ref{tab:depth_demon_test}. We outperform previous representative methods (\eg, DPSNet) by a large margin. Compared with IIB~\cite{yifan2022input} which injects the geometric inductive bias directly to the input of the Transformer, our performance is similar but our method is more lightweight and faster, which demonstrates that a better modeling of the geometric inductive bias enables the problem to be solved more efficiently.

\section{Limitation and Discussion}

Our work has several limitations. First, our method still has room for further improvement in unmatched regions. As shown in Table~\ref{tab:flow_sintel_test}, the performance in matched regions on the Sintel dataset is already very accurate (with an end-point-error of 0.34 pixels on the clean split and 1.10 pixels on the final split). However, the error in unmatched regions is considerably larger, which deserves further investigation in future. Second, we resort to RAFT's iterative refinement architecture as a post-processing step to further improve our performance. We believe more lightweight and effective approach would be applicable which we consider as interesting future work. Third, our full model is still not able to achieve real-time inference speed. Further improvements are necessary to enable applications with real-time requirements (20 FPS or more). Finally, in this paper, we have demonstrated the applicability of our method to multiple 2-frame tasks. We consider the extension of our approach to the multi-view scenario as an interesting future direction.

Our unified model might also shed some light on training a single model to do all three tasks \textit{simultaneously}. In this paper, we haven’t shown such experiments yet. There are also additional challenges to resolve (\eg, how to balance different tasks in the joint training process). Besides, to train such a unified model, one could also explore recent unsupervised pretraining approaches (\eg, masked autoencoders~\cite{he2022masked}) to learn general feature representations for matching. We believe that our work may serve as a fruitful basis for further research in this area.

\section{Conclusion}

We have presented a unified formulation and model for three different motion and 3D perception tasks: optical flow, rectified stereo matching and unrectified stereo depth estimation. We demonstrated that all three tasks can be solved with a unified model by formulating them as a unified dense correspondence matching problem. This allows to reduce the problem to learning high-quality discriminative features for matching, for which we use a Transformer, in particular exploiting its cross-attention mechanism to integrate information from the other view. One unique benefit of our unified model is that it naturally enables cross-task transfer since all the learnable parameters are exactly the same. Our final model achieves state-of-the-art or highly competitive performance on 10 popular flow/stereo/depth datasets, while being simpler and more efficient in terms of model design and inference speed. 

A key result of this paper is that features aggregated via a Transformer from both input images are stronger and contain more discriminative correspondence information, which enables to greatly simplify existing motion and depth estimation pipelines, while achieving improved performance. We hope our findings can be useful for more dense correspondence and multi-view perception tasks.

\vspace{-10pt}
\ifCLASSOPTIONcompsoc
  \section*{Acknowledgments}
\else
  \section*{Acknowledgment}
\fi

Andreas Geiger was supported by the ERC Starting Grant LEGO-3D (850533) and the DFG EXC number 2064/1 - project number 390727645. Jing Zhang and Dacheng Tao were supported by ARC FL170100117.

\ifCLASSOPTIONcaptionsoff
  \newpage
\fi

\bibliographystyle{IEEEtran}
\bibliography{IEEEabrv,bib}

\end{document}